\documentclass{article}

\usepackage[preprint]{tmlr} 

\usepackage[colorlinks=true,linkcolor=blue,citecolor=blue,urlcolor=blue]{hyperref}

\usepackage{amsmath,amsfonts}
\usepackage{graphicx}
\graphicspath{{Images/}}
\usepackage{amssymb}
\usepackage{bbm}

\usepackage{booktabs}
\usepackage{threeparttable}
\usepackage{tabularx}
\usepackage{multirow}
\usepackage{array}
\usepackage{float}
\usepackage{adjustbox}
\usepackage{subcaption}
\usepackage[table]{xcolor}
\usepackage{enumitem}

\usepackage{amsmath,amsfonts,bm}

\def\eqref#1{equation~\ref{#1}}

\def\1{\bm{1}}

\DeclareMathAlphabet{\mathsfit}{\encodingdefault}{\sfdefault}{m}{sl}
\SetMathAlphabet{\mathsfit}{bold}{\encodingdefault}{\sfdefault}{bx}{n}

\newcommand{\openreview}{\url{https://openreview.net/forum?id=QQj5zaYQED}}

\begin{document}

\title{MTL-MHOL: Predicting Online Conversions under Delayed Feedback and Data Sparsity}

\author{\name Mario Tejera Vicente \email mario.tejera@icloud.com \\
        \addr Erasmus School of Economics\\
        Erasmus University Rotterdam\\
        Rotterdam, 3062 PH, Netherlands
        \AND
        \name Eva Hagen\textsuperscript{*} \email evahagen6@gmail.com \\
        \addr Erasmus School of Economics\\
        Erasmus University Rotterdam\\
        Rotterdam, 3062 PH, Netherlands
        \AND
        \name Emma van Breukelen\textsuperscript{*} \email emmavanbreukelen@outlook.com \\
        \addr Erasmus School of Economics\\
        Erasmus University Rotterdam\\
        Rotterdam, 3062 PH, Netherlands
        \AND
        \name Quinten van de Vijver\textsuperscript{*} \email quintenvdvijver02@gmail.com \\
        \addr Erasmus School of Economics\\
        Erasmus University Rotterdam\\
        Rotterdam, 3062 PH, Netherlands
        \AND
        \name Kathrin Gruber \email gruber@ese.eur.nl \\
        \addr Erasmus School of Economics\\
        Erasmus University Rotterdam\\
        Rotterdam, 3062 PH, Netherlands}

\maketitle

\vspace{-0.5em}
{\small\textsuperscript{*}The second, third and fourth authors contributed equally to this paper.}
\vspace{0.5em}

\begin{abstract}
\looseness=-1 This paper proposes the model-agnostic Multi-Task Learning Multi-Head Online Learning (MTL-MHOL) framework for conversion rate (CVR) prediction. 
Existing approaches typically address key challenges in CVR prediction, such as delayed feedback and data sparsity, in isolation or lack flexibility and practical applicability. 
MTL-MHOL adopts a time bucketing approach to account for delayed feedback and combines it with multi-task learning of an auxiliary task to mitigate data sparsity. 
We evaluate the framework on proprietary datasets from a private company and on a public dataset from Criteo. MTL-MHOL matches or outperforms all benchmark models in terms of Negative Log-Likelihood (NLL) and Relative Cross Entropy (RCE), and it correctly captures temporal trends in the data using an MLP backbone, while maintaining strong performance with a DeepFM backbone.
In particular, MTL-MHOL matches the performance of the advanced delayed-feedback method FSIW, outperforms the entire-space approach ESMM by 21.5\% in RCE, and achieves up to 81\% lift in RCE compared to the best-performing classical benchmark. The code is freely available at \url{https://github.com/emmavanbreukelen/MTL-MHOL}.


\end{abstract}



\section{Introduction}
\label{intro}

\looseness=-1
In online advertising, conversion rate (CVR) prediction is central to modeling user behavior. It is defined as estimating the probability that a user who interacts with a company's online content performs a desired action, such as making a click or a purchase.
A wide range of models have been developed to address three recurring challenges: delayed feedback, data sparsity and selection bias \citep{chapelle2014modeling, ktena2019addressing, gao2022multihead,lee2012estimating, zhang2014multi, ji2016probabilistic, xia2020new, liu2023task, dishi2025practical, ma2018entire, wang2020delayed}.
Yet these challenges are typically addressed within separate modeling paradigms. Multi-head approaches \citep[e.g.,][]{gao2022multihead} provide an effective solution to delayed feedback through time bucketing, while ESMM-style models \citep{ma2018entire} address data sparsity and selection bias via entire-space multi-task learning. The few attempts to bridge these paradigms \citep{wang2020delayed, Zhao2023} lack the architectural flexibility required for continuous online deployment.


\looseness=-1
The delayed feedback problem results from conversions that often occur long after the corresponding impression or click, leading to false negatives at training time. This is especially relevant in online settings where models update continuously on either fresh but incomplete labels, or matured but stale labels \citep{chapelle2014modeling}. The data sparsity problem refers to conversions being rare events relative to impressions or clicks, making the corresponding datasets highly imbalanced with low signal \citep{ma2018entire}. Moreover, training is typically performed only on clicked impressions, while inference utilizes the entire space of impressions. This leads to biased estimates and harms generalizability  \citep{zadrozny2004learning}.

A substantial body of work has focused on mitigating data sparsity and sample selection bias through multi-task learning and entire-space modeling. 
The Entire Space Multi-task Model \citep{ma2018entire} addresses this by jointly learning click-through rate (CTR) and post-click CVR  over the full impression space, such that the abundant click signal regularizes the sparse conversion task.
This entire-space formulation has since been extended through post-click behavior decomposition and counterfactual multi-task learning \citep{wen2020entire, wang2022escm2}. More recent causal multi-task approaches explicitly correct for exposure and logging biases while preserving shared representations \citep{wang2022escm2, zhu2023dcmt}. However, none of these approaches model the temporal structure of conversion delays, leaving them unable to exploit partially mature observations or predict when a conversion is likely to occur.
    
Likewise, a dedicated line of work tackles delayed feedback directly. \citet{chapelle2014modeling} introduces a survival-style model to label non-converted clicks as negative or too-early-to-tell, reducing label bias under delay. \citet{ktena2019addressing} propose a delayed feedback loss and label-update mechanism for continuous neural network training. \citet{yasui2020} introduce an importance weighting algorithm to reduce the increased negative rate in the loss function. 
However, all of these models use a single time window, which limits their ability to exploit very recent data while still accounting for long-tail conversions. \citet{gao2022multihead} address this with a Multi-Head Online Learning (MHOL) framework that partitions conversion delays into mutually exclusive time windows, each handled by a dedicated head. Yet MHOL operates in a single-task, click-space setting and cannot leverage auxiliary signals or non-clicked impressions to alleviate sparsity and selection bias. This leaves an opportunity to integrate MHOL with an entire-space multi-task framework.

\enlargethispage{1em}
\subsection{Our Contributions}

This paper proposes the flexible and model-agnostic Multi-Task Learning Multi-Head Online Learning (MTL-MHOL) framework which jointly mitigates delayed feedback and data sparsity in CVR prediction. The architecture consists of two tasks: the primary task focuses on CVR prediction through delayed feedback modeling, the auxiliary task enhances CVR prediction through the use of an engagement measure. 
We present the framework in the context of a Multi-Layer Perceptron (MLP) as the base architecture and illustrate the model-agnostic behavior of our approach with a DeepFM-based implementation.

\looseness=-1
For the primary task, we adapt the multi-head model of \citet{gao2022multihead} to split the full conversion period into shorter time windows, each with its own model trained on the most recent mature data available. A survival-based formulation then aggregates the window-specific outputs into a single CVR estimate.
This way, our framework enables continuous training and updating based on the most recent observations. Moreover, by defining the primary CVR task over the entire impression space rather than conditioning on clicks, we reduce sample selection bias caused by training on a selected subset of the entire data.

\looseness=-1 The auxiliary task is trained jointly with the primary task via hard parameter sharing \citep{caruana1993multitask}, providing additional learning signals in settings where conversion data is sparse. 
With very few converting observations, the primary task alone offers little information to learn from; the auxiliary task compensates for this during training, which yields more robust shared representations and, in turn, more accurate CVR predictions.  
For continuous auxiliary targets, we introduce a multi-quantile regression head with a soft monotonicity penalty to obtain a richer supervisory signal.

We train and evaluate our model on data provided by a private marketing company and on a public dataset provided by the advertising company Criteo.
The company datasets consist of samples of user sessions with associated conversion outcomes along with other relevant variables that may influence customer behavior collected by the company. The Criteo data contains impression, click and conversion data widely used for benchmarking CVR-related models. The models are evaluated on Negative Log-Likelihood (NLL) and Relative Cross Entropy (RCE) in a rolling window cross-validation setting.

\looseness=-1 Our MTL-MHOL model matches the performance of neural delayed-feedback (FSIW), and outperforms classical single-head baselines (LR and RF trained on fully matured labels) and entire-space baselines (ESMM), as well as all ablation variants (MLP, MTL, and MHOL) for the MLP backbone. 
Moreover, strong performance under both MLP and DeepFM backbones confirms that MTL-MHOL is model-agnostic within the
neural network family. The model is also able to distinguish conversions that are more likely to occur soon and those likely to occur later, regardless of different temporal patterns in different datasets. 

In sum, we make the following contributions to CVR prediction in online advertising environments:
\begin{itemize}[leftmargin=*, itemsep=1pt]

    \item We propose MTL-MHOL, a framework that integrates bucketed multi-head online learning for delayed feedback with multi-task learning for data sparsity into a unified architecture suitable for continuous online deployment, and formalize a risk-set indicator that restricts each head's training population to at-risk observations --- necessary for the survival aggregation to remain a valid probability decomposition.
    
    \item We introduce a hard parameter-sharing auxiliary task design supporting either a binary or a continuous engagement target, where the continuous case uses a multi-quantile regression head with a soft monotonicity penalty --- providing richer supervision than a single point estimate and novel relative to both MHOL and ESMM-family work.

    \item We show that MTL-MHOL produces well-calibrated, delay-aware predictions across time buckets, correctly capturing dataset-specific conversion timing patterns and enabling prediction of not only whether but also when a conversion occurs.


    \item We demonstrate through empirical analysis on proprietary company datasets and a public Criteo dataset that MTL-MHOL matches the performance of the advanced delayed-feedback method FSIW, outperforms the entire-space approach ESMM, the classical baselines (LR and RF), and all ablation variants (MLP, MHOL, and MTL), while maintaining strong performance across both MLP and DeepFM backbones, demonstrating its model-agnosticity within the neural network family.
\end{itemize}

\begin{figure}[t!]
    \centering
    \includegraphics[width=0.4\linewidth]{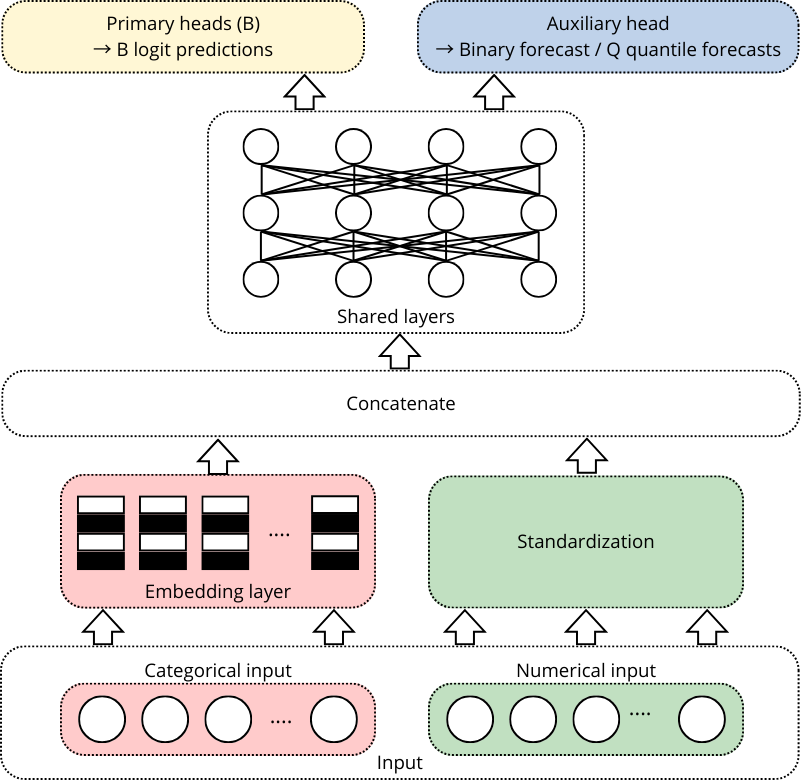}
    \caption{\looseness=-1 MTL-MHOL combines multi-head online learning for delayed feedback modeling with multi-task learning to incorporate auxiliary engagement signals for sparse CVR prediction, here illustrated with an MLP backbone.
    Categorical variables are embedded in a fixed-dimensional space. Standardized numerical values are used as one-dimensional representations and concatenated with the embeddings of categorical variables. The outputs can either be single output layers or specialized MLP towers.}
\label{fig:visualisation_model}
\end{figure}

\section{Methodology} \label{Sect: methodology}
 
\looseness=-1
This paper proposes Multi-Task Learning Multi-Head Online Learning (MTL-MHOL) for CVR prediction. 
The framework builds on the model-agnostic MHOL blueprint of \citet{gao2022multihead} with a multi-task setting in order to jointly tackle delayed feedback and data sparsity. The MHOL design balances training data staleness and the false-negative rate in predictions, by targeting CVR prediction across mutually exclusive and collectively exhaustive delay horizons. This enables the model to incorporate information as soon as labels have matured for the shortest delay horizon, thereby ensuring that only fully matured labels are used for training. We extend this principle to a multi-task setting to enable the delay-specific heads of the MHOL component to be jointly trained with a binary or continuous auxiliary  
dense-signal task, and restrict each primary head's training population to at-risk observations. 
A visualization of the model architecture is provided in \autoref{fig:visualisation_model}.

\subsection{Workhorse Models} \label{sec:workhorse}

\looseness=-1 
We present our framework in the context of a Multi-Layer Perceptron (MLP) as the base architecture. MLPs are simple feedforward networks with fully connected layers and serve as a natural and widely adopted choice for analysis in this paper. Yet, specific modeling needs can be met by replacing the MLP with any differentiable architecture.
For example, recurrent architectures such as Long Short-Term Memory \citep[LSTM,][]{hochreiter1997long} or Gated Recurrent Unit \citep[GRU,][]{cho2014learning} models may be employed when temporal dependencies are central, whereas models such as Deep Factorization Machines \citep[DeepFM,][]{Guo2017DeepFM} may be preferable for capturing low-degree feature interactions.\footnote{Non-backpropagation backbones (e.g.,~kernel methods, tree ensembles) are also viable in principle \citep{Caruana1997}, though largely unexplored in this setting.} 
The framework imposes no architectural constraints beyond differentiability; empirical demonstration in this paper is limited to MLP and DeepFM, which together represent a broad-capacity feedforward baseline and an interaction-aware architecture.

\subsubsection{MLP} \label{sec:mlp}

In CVR prediction, data frames commonly hold categorical and numerical features, which can be preprocessed in several different ways depending on the specific characteristics. We discuss our preprocessing pipeline, including the imputation of missing values, in Appendix~\ref{Appendix: private data} and \ref{Appendix: criteo data}. Categorical variables are represented using embedding layers that map each category to a dense vector representation, while numerical variables are used in their one-dimensional standardized representation per numerical field. These categorical and numerical representations are then concatenated to produce a single input vector per sample, which is passed to the MLP architecture.
    
Subsequently, the representation is processed by a shared trunk network that culminates in branches of task-specific towers. For the primary multi-head CVR prediction task, the model outputs one node per delay-specific bin, each passed through a sigmoid activation function. For the distributional auxiliary task, the model produces as many logits as quantiles to be estimated, using a linear output layer. In case of a binary auxiliary task, the corresponding tower outputs one logit that is fed into a sigmoid function.

\subsubsection{DeepFM}

To illustrate the adaptability of the framework, we complement our MLP implementation with a DeepFM model \citep{Guo2017DeepFM}. That is the combination of the deep learning characteristics of an MLP together with the wide learning achieved by a Factorization Machine \citep[FM,][]{rendle2010factorization}. A detailed formulation of the used implementation can be found in \autoref{app:deepFM}. The model learns relevant low- and high-order interactions in parallel through simultaneously training the FM and Deep Neural Network components based on a joint embedding layer. For the multi-task implementation, we follow the guidelines of \citet{yildirim2021bideepfm}, who propose a biDeepFM framework for reciprocal recruitment recommender systems. We adapt this formulation slightly, creating a framework that is able to allocate single as well as multiple output heads for either one of the two tasks. We note that the DeepFM framework is foremost suitable in cases of scarce and shallow interaction terms being particularly informative.

\subsection{Multi-Head CVR Prediction}

\looseness -1 The delayed feedback problem arises from labels not being observed immediately after an event, leading to stale supervision when restricting training to fully matured data and increased false negatives when shorter waiting times are used. However, in online learning environments --- settings with continuous label-driven model updates as new observations mature \citep{ktena2019addressing} --- models must be continuously updated with recent data to capture emerging trends and behavioral shifts. We therefore tackle delayed feedback with an MHOL framework.

\subsubsection{Delayed Feedback Modeling} \label{sec: delayed feedback modeling}

\looseness=-1
Instead of predicting a single conversion probability over the entire time horizon, we break the problem down into smaller subproblems. Specifically, the full time horizon $H$, during which a conversion can occur, is divided into $B$ separate buckets such that each bucket $b \in \{1,2,\dots,B\}$ corresponds to a time window $w_b = [t_{b-1}, t_{b})$ up to $w_B = [t_{B-1}, t_{B}]$ within $H = [t_0, t_B]$. Here, $t_b$ denotes the end timestamp of bucket $b$, whereas $t_0$ refers to the log-timestamp of the observation. A dedicated sub-model is trained for each bucket, forming the heads of the multi-head architecture. All heads share the same input features and trunk parameters but differ in their targets, namely predicting CVR within their own time windows.

For every observation $i$, we construct a vector $\mathbf{y}_i$ with $B$ elements $[y_{i1},\dots,y_{iB}]$, where each element equals $1$ if a conversion happened within the corresponding time window $w_b$, and $0$ otherwise. This formulation allows the model to incorporate supervision as soon as the earliest time window $w_1$ has matured. 
    
\begin{figure}
    \centering
        \includegraphics[width=0.53\linewidth, height=0.15\textheight]{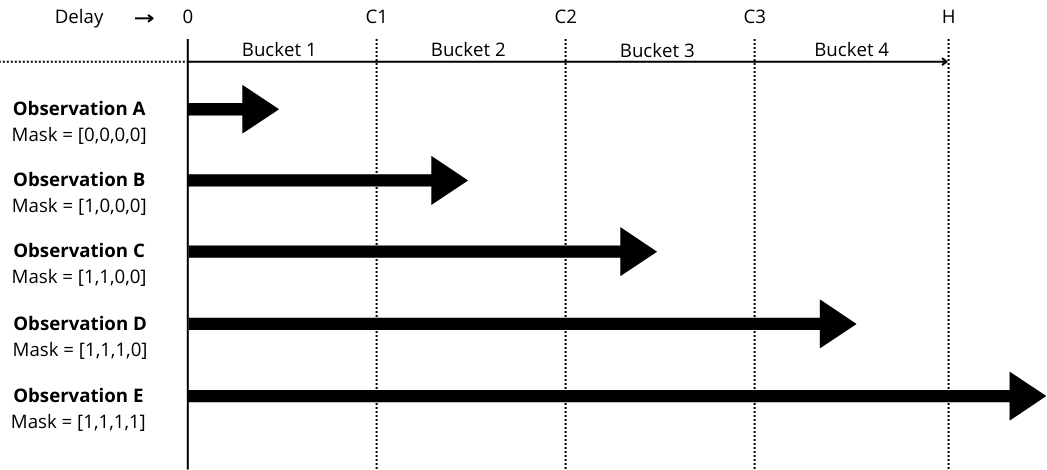}
    \caption{The five masking scenarios for a non-converter across $B=4$ sequential delay buckets. Arrows indicate elapsed time from observation start to the current training cutoff; $C_X$ denotes bucket cutoff $X \in \{1,2,3\}$; $H$ is the maximum conversion delay.}
    \label{Masking visualization}
    \vskip -0.1in
\end{figure}

However, not all bucket labels are observed at a given cutoff time $\varphi$. We therefore employ a maturity masking approach to control which time windows contribute to training at $\varphi$. We define a maturity mask $\mathbf{m}_i = [m_{i1}, \dots, m_{iB}]'$, where
\begin{equation} \label{eq:mask}
    m_{ib} = \begin{cases}
        1, & \text{if } t_b \leq \varphi, \\
        0, & \text{otherwise,}
    \end{cases}
\end{equation}
\looseness=-1 indicating whether bucket $b$ has fully elapsed by cutoff $\varphi$.
For each fold, $\varphi$ is set to the end of the training data. The observation's age is computed as the difference between the cutoff time and the observation's start timestamp. If no conversion occurs, a time window is marked as mature only after its full duration has elapsed. If a conversion occurs before the cutoff time, all windows are marked as mature, since the true bucket-level outcome is then fully determined for every $b$. This ensures that negative labels are assigned only after the full time window has elapsed and prevents premature negative classification. 
Maturity alone, however, only certifies that a label is known with certainty: it does not by itself determine whether observation $i$ should be used to train a given head, since each head is trained on a bucket-specific risk set that we make explicit in Section~\ref{sec:pred}. Figure~\ref{Masking visualization} illustrates the five masking scenarios that can arise for a non-converter across $B=4$ buckets.

\subsubsection{Prediction} \label{sec:pred}

The multi-head architecture outputs $B$~logits per observation, one for each delay bucket. 
Let $y_{ib} \in \{0,1\}$ indicate whether observation~$i$ converts in bucket~$b$, and let $\mathcal{X}_i$ denote the information available at the start of observation~$i$. The sigmoid $\sigma(\cdot)$ maps the logit $l_{ib}$ to the hazard probability
\begin{equation} \label{eq:hazard}
    h_{ib} = \sigma(l_{ib})
    =
    \Pr\!\left(y_{ib} = 1 \mid y_{i1} = \dots = y_{i(b-1)} = 0,\,
    \mathcal{X}_i\right),
\end{equation}
that is, the probability of a conversion in bucket~$b$ given no conversion in any earlier bucket. At most one bucket per observation can equal one, so the conditioning on no prior conversion is natural. 

This implies that head $b$ should only be trained on observations still at risk, as formalized by the indicator
\begin{equation}
    r_{ib} = \mathbbm{1}\!\left\{\textstyle\sum_{b'=1}^{b-1}
    y_{ib'} = 0\right\},
    \label{eq:riskset}
\end{equation}
where $r_{ib} = 1$ if and only if observation $i$ has not converted before bucket $b$. By construction $r_{i1} = 1$ for all $i$, and once $y_{ib'} = 1$ for some $b' < b$, $r_{ib} = 0$ for all subsequent buckets. 
Note that $r_{ib}$ is distinct from the maturity mask $m_{ib}$ of Equation~\ref{eq:mask}: $m_{ib}$ certifies that bucket $b$'s label is fully known by cutoff $\varphi$, while $r_{ib}$ certifies that observation $i$ belongs to the population $h_{ib}$ describes. Both are required jointly for an observation to contribute to head $b$'s training loss, as formalized in Section~\ref{sec:loss}. The per-bucket hazards also provide directly interpretable online forecasts that can be updated as each bucket matures.

\looseness=-1
To aggregate the bucket-specific hazards, let $S_i$ denote the survival probability of observation~$i$, that is, the probability of no conversion in any of the $B$~buckets:
\begin{equation}\label{eq:survival}
    S_i = \prod_{b=1}^{B} (1 - h_{ib}).
\end{equation}
Since each hazard already conditions on survival up to its bucket, no further independence assumption is required, and the overall predicted conversion probability follows as the complement of the survival probability:
\begin{equation} \label{eq:phat}
    \hat{p}_i
    =
    \Pr(y_i = 1 \mid \mathcal{X}_i)
    =
    1 - S_i
    =
    1 - \prod_{b=1}^{B} (1 - h_{ib}).
\end{equation}
For evaluation in hindsight, we compare $\hat{p}_i$ against the realized label $y_i = \mathbbm{1}\!\left(\sum_{b=1}^B y_{ib} \neq 0\right)$, which is observed only once the full observation window has elapsed.

\subsection{Auxiliary Task}\label{Methodology: auxiliary task}

The multi-head architecture gains temporal resolution at a cost: splitting the conversion horizon across $B$ buckets dilutes an already sparse signal, risking instability and overfitting within each head. We counter this dilution by introducing an auxiliary task trained jointly with the primary CVR task via hard parameter-sharing \citep{caruana1993multitask}, which lets the shared trunk learn from a denser supervision signal without altering the primary objective. A review of the benefits and risks of MTL frameworks is provided in Appendix~\ref{app:aux}.

Depending on the dataset, the auxiliary target is either binary or continuous. For continuous targets we adopt a multi-quantile regression head for two reasons: predicting multiple quantiles yields a richer representation of the conditional distribution than a single point estimate \citep{Ruder2017}, and quantile regression imposes no parametric assumptions, making it robust to the skewness and heavy tails common in engagement variables.

\subsection{Loss Functions} \label{sec:loss}

\paragraph{Primary task loss.} Each bucket head is trained with a masked binary cross-entropy loss restricted to its risk set. Let $y_{ib} \in \{0,1\}$, $m_{ib} \in \{0,1\}$, and $r_{ib} \in \{0,1\}$ denote the conversion label, maturity mask (Equation~\ref{eq:mask}), and risk-set indicator (Equation~\ref{eq:riskset}) for observation $i$ in bucket $b$, and let $\hat{h}_{ib} \in (0,1)$ be the predicted hazard. Defining the joint eligibility weight $\omega_{ib} = m_{ib}\,r_{ib}$, the loss for bucket $b$ is
\begin{equation}
    \mathcal{L}_{CE_b} = -\frac{1}{\sum_{i=1}^N \omega_{ib}}
    \sum_{i=1}^N \omega_{ib}
    \Big[y_{ib} \log(\hat{h}_{ib}) + (1-y_{ib})\log(1-\hat{h}_{ib})\Big],
    \label{Lbce}
\end{equation}
the average negative log-likelihood over all mature, at-risk observations in bucket $b$. The risk-set restriction ($r_{ib} = 0$ for prior converters) is essential: without it, each head converges toward the marginal $\Pr(y_{ib}=1 \mid \mathcal{X}_i)$ rather than the conditional hazard of Equation~\ref{eq:hazard}, invalidating the survival decomposition in Equations~\ref{eq:survival}--\ref{eq:phat}.

\paragraph{Auxiliary task loss.} For a binary auxiliary target, we reuse the loss in Equation~\ref{Lbce} with two simplifications: there is only one head $B = 1$, and the maturity mask is replaced by an auxiliary mask $a_i \in \{0,1\}$. The mask is 1 by default and set to 0 only when a maturity timestamp is available and the auxiliary outcome is not yet observable at cutoff $\varphi$. The resulting loss is
\begin{equation}\label{Lbce_aux}
    \mathcal{L}_{\text{aux}}^{\text{binary}} = - \frac{1}{\sum_{i=1}^{N} a_{i}}
    \sum_{i=1}^{N} a_{i}
    \left[
    y^{\text{aux}}_{i} \log(\hat{h}_{i}^{\text{aux}}) +
    (1 - y^{\text{aux}}_{i})\log(1 - \hat{h}_{i}^{\text{aux}})
    \right],
\end{equation}
where $\hat{h}_{i}^{\text{aux}}$ is the predicted probability that $y_i^{\text{aux}} = 1$.
    
\looseness=-1 
For a continuous auxiliary target, a single point estimate would discard distributional information that could otherwise enrich the shared representation. We therefore train the auxiliary head to predict $Q$ conditional quantiles simultaneously via a pinball loss. Let $\hat{y}^{\text{aux}}_{\tau_q,i}$ be the predicted conditional quantile at level $\tau_q \in (0,1)$, $q = 1,\dots,Q$. The pinball loss is
\begin{equation}
    \mathcal{L}_{\text{pinball}} =
    \frac{1}{\sum_{i=1}^{N} (a_{i} \cdot Q)} \sum_{i=1}^{N} a_{i} \sum_{q=1}^{Q}
    \max \Big( \tau_q \, (y^{\text{aux}}_{i} - \hat{y}^{\text{aux}}_{\tau_q, i}),\,
    (\tau_q - 1)(y^{\text{aux}}_{i} - \hat{y}^{\text{aux}}_{\tau_q, i}) \Big),
    \label{eq_pinball}
\end{equation}
which applies an asymmetric penalty to target each conditional quantile of the auxiliary distribution. To ensure the predicted quantiles form a valid conditional distribution, we add the soft monotonicity penalty of \citet{decke2024efficient} to penalize any violation of the ordering $\hat{y}^{\text{aux}}_{\tau_q,i} \leq \hat{y}^{\text{aux}}_{\tau_{q+1},i}$:
%
$
    \mathcal{P} =
    1/\big(\sum_{i=1}^{N} a_{i} \cdot (Q-1)\big)
    \sum_{i=1}^{N} a_{i} \sum_{q=1}^{Q-1}
    \max\!\big(0,\, \hat{y}^{\text{aux}}_{\tau_q, i} -
    \hat{y}^{\text{aux}}_{\tau_{q+1}, i}\big)
$.

The combined auxiliary loss is then
\begin{equation}
    \mathcal{L}_{\text{aux}}^{\text{distr}}
    = \mathcal{L}_{\text{pinball}}
    + \gamma \, \mathcal{P},
    \label{eq_auxloss1}
\end{equation}
where $\gamma \geq 0$ controls the strength of the monotonicity penalty.

\paragraph{Joint loss.} The overall training objective is
\begin{equation} \label{eq:Loss}
    \mathcal{L} \;=\; -\,\frac{1}{\sum_{b=1}^{B}\sum_{i=1}^{N}\omega_{ib}}
    \sum_{b=1}^{B}\sum_{i=1}^{N}\omega_{ib}\Big[\,y_{ib}\log\big(\hat{h}_{ib}\big)
    + \big(1-y_{ib}\big)\log\big(1-\hat{h}_{ib}\big)\Big]
    \;+\; \lambda_{\mathrm{aux}}\,\mathcal{L}_{\mathrm{aux}}.,
\end{equation}
where $\lambda_{\text{aux}}$ scales the auxiliary contribution and $\mathcal{L}_{\text{aux}}$ is either $\mathcal{L}_{\text{aux}}^{\text{binary}}$ or $\mathcal{L}_{\text{aux}}^{\text{distr}}$. The primary loss thus remains the dominant training signal, with the auxiliary term acting as a data-adaptive regularizer. Empirically, $\lambda_{\text{aux}}$ adds only marginal overhead to hyperparameter tuning. A sensitivity analysis on the Criteo data  shows that for $\lambda_{\text{aux}} = \{0.1, 0.5, 0.9\}$, relative RCE varies by $0.034$ for MTL-MHOL and $0.049$ for MTL (see Appendix~\ref{app:sens_analysis} for a detailed discussion).

\section{Experimental Setup} \label{Sect: experimental setup}

The experiments evaluate MTL-MHOL along four dimensions: benchmark comparison, component-wise ablation, bucket-level temporal accuracy, and backbone generalizability. Each dimension corresponds to one contribution and is addressed in Sections~\ref{sec:ablation}--\ref{sec:model-agnostic}, using NLL and RCE under a rolling window cross-validation design (see Appendix~\ref{Appendix HPO}).

\subsection{Data}

We evaluate MTL-MHOL on two datasets: impression-level data from Criteo and session-level data provided by a private marketing company. Summary statistics for all datasets are reported in Table~\ref{tab:summarystats} in Appendix~\ref{Appendix Data}. The difference in observation unit allows us to assess the framework across complementary data regimes.

\paragraph{Criteo Data.} We use the Attribution Modeling for Bidding dataset \citep{criteoref}, a widely adopted benchmark in CVR prediction research, repurposed for conversion prediction given an impression. 
Downsampling preserves the original conversion rate (4.90\% full dataset, 4.93\% after downsampling) and the relative delay distribution within the $H = 12$ day window.
Downsampling, preprocessing, and feature engineering are detailed in Appendix~\ref{Appendix: criteo data}.

\paragraph{Private Data.} The private dataset covers two customers of a marketing company. Full preprocessing details are provided in Appendix~\ref{Appendix: private data}.

\subsection{Bucket Construction} 

\looseness=-1 Following Section~\ref{sec: delayed feedback modeling}, the maximum delay window $H$ is divided into four buckets, each assigned a binary conversion label. Bucket thresholds are selected from the empirical delay distribution to ensure sufficient training observations per bucket while remaining interpretable. At training time, the cutoff $\varphi$ is set to the end of the training window, so that $m_{ib} = 1$ only for buckets whose upper boundary $t_b$ satisfies $t_b \leq \varphi$. The resulting bucket boundaries and conversion distributions are reported in Appendix~\ref{Appendix Bucket construction}; delay distribution plots for both datasets are in Appendix~\ref{Appendix Descriptive Stats} and \ref{Appendix Criteo descriptive stats}.

\subsection{Auxiliary Task}

\paragraph{Criteo Data.} \looseness=-1 We use the binary click indicator as the auxiliary target. Since clicks precede conversions and are observed far more frequently, they provide a denser supervision signal that alleviates the sparsity of the primary CVR task.
The strong empirical association between clicks and conversions suggests that both tasks share underlying user behavior patterns such that representations learned from the abundant click signal transfer naturally to the rare conversion task through the shared trunk.

\begin{figure}[t!]
\vskip -0.9em
    \centering
    \begin{subfigure}{0.49\textwidth}
    \centering
    \includegraphics[width=0.93\linewidth, height=0.18\textheight]{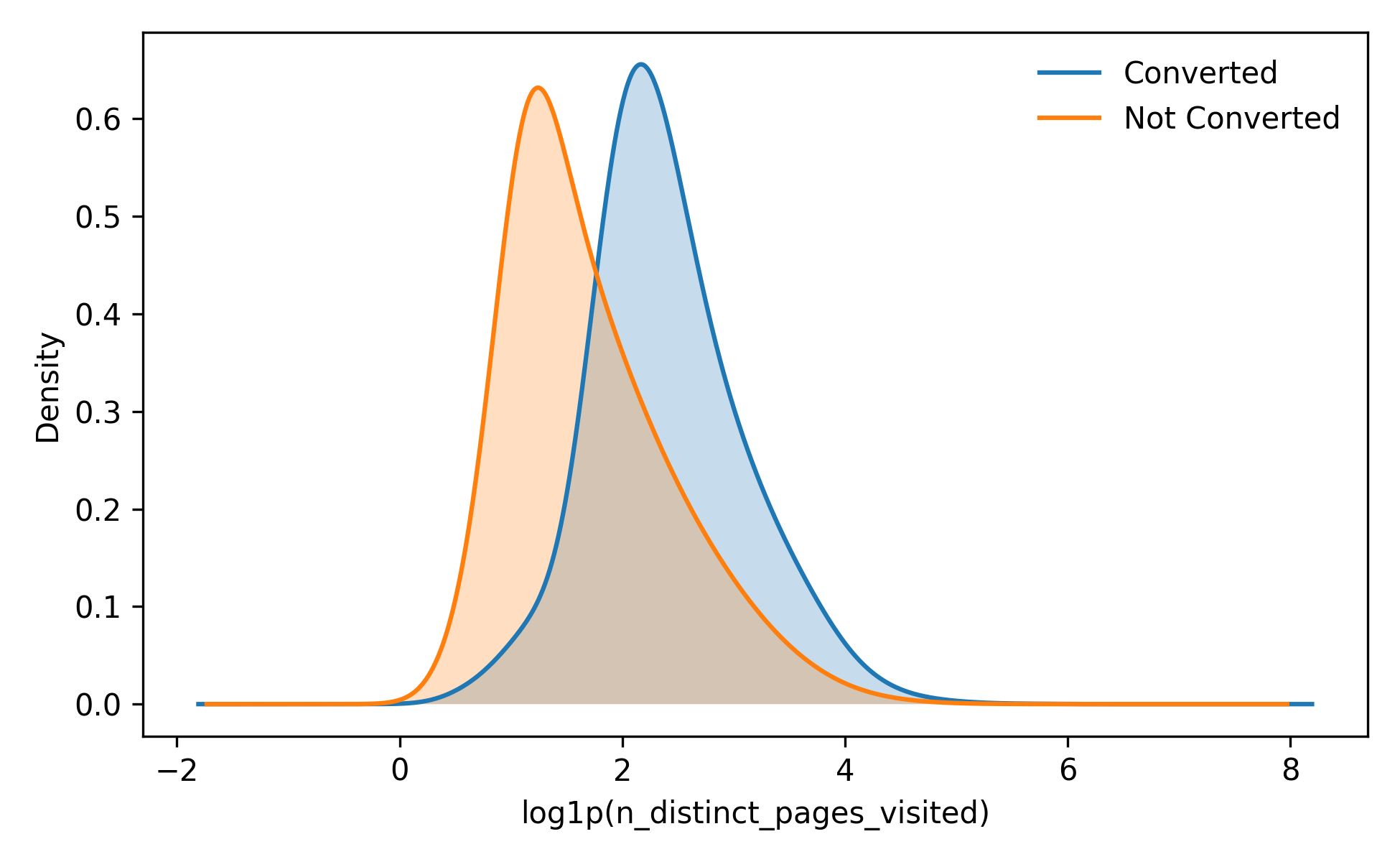}
    \caption{Customer~1}
\end{subfigure}
\hfill
\begin{subfigure}{0.49\textwidth}
    \centering
    \includegraphics[width=0.93\linewidth, height=0.18\textheight]{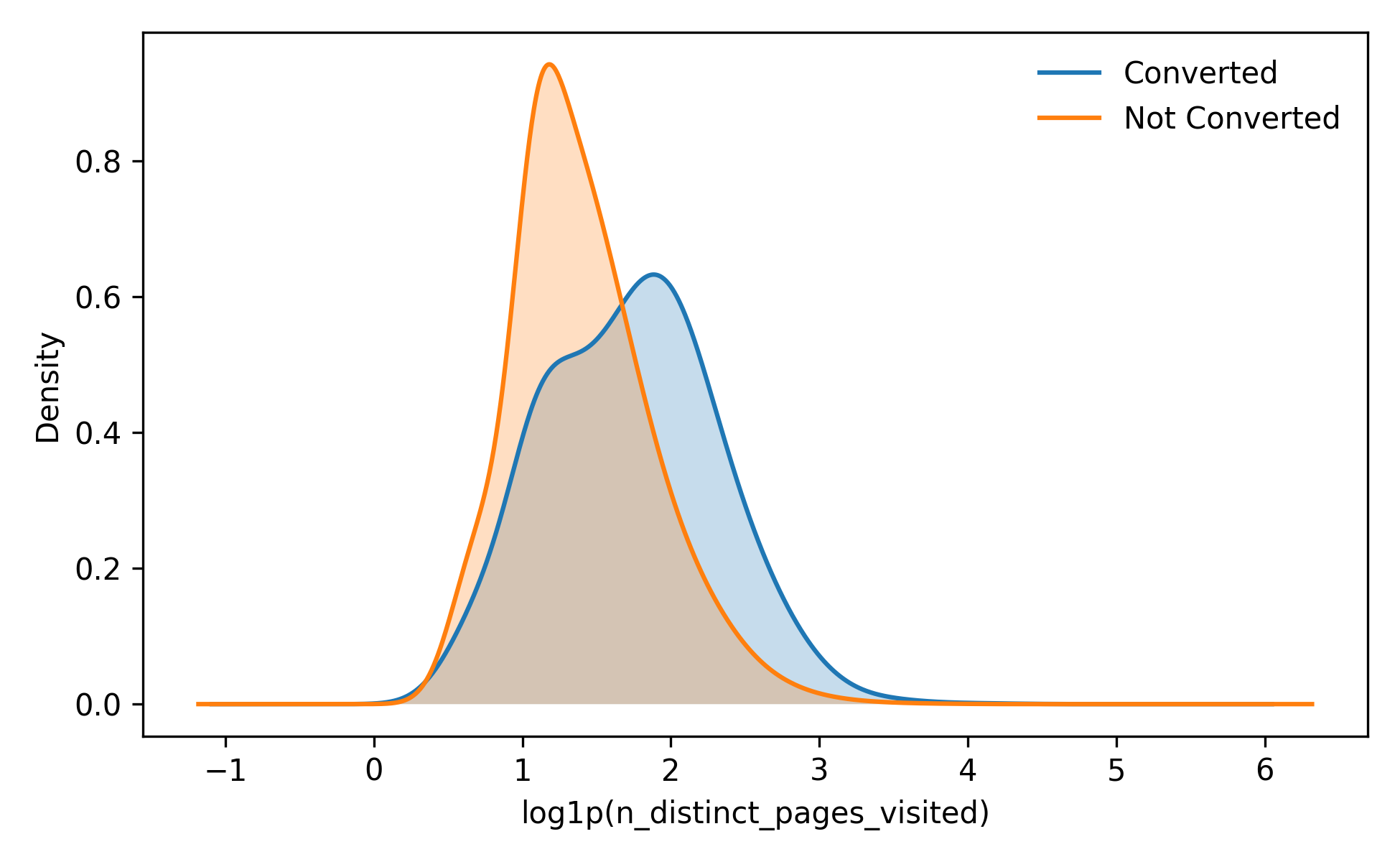}
    \caption{Customer~2}
\end{subfigure}
    \caption{Kernel density of the log-transformed number of distinct pages visited by conversion status for Customers~1 and 2 after removing short sessions.}
    \label{kde_n_distinct_main}
\vskip -0.7em
\end{figure}

\paragraph{Private Data.} Here, we use the number of distinct pages visited per session as the auxiliary target to reflect the intensity of user engagement.
The variable is observed for all sessions, making it far more frequent than conversion events. 
Figure~\ref{kde_n_distinct_main} shows that converted sessions involve more distinct page visits on average than non-converted ones. The latter are heavily concentrated at low page counts, confirming that users who browse more pages are more likely to be actively evaluating products and thus more likely to convert, suggesting shared underlying representations. The variable is nevertheless heavily right-skewed with high variance, so the auxiliary target is log-transformed prior to modeling to reduce skewness and improve numerical stability. 
The auxiliary head predicts multiple conditional quantiles per session, with the number of quantiles treated as a tunable hyperparameter (Appendix~\ref{Appendix HPO}). Candidate quantile levels span the lower, median, and upper regions of the conditional distribution.
Appendix~\ref{app:aux_target_type} explores the possibility of using a binary auxiliary task instead and discusses the findings for Customer~2.

\subsection{Architectural Choices}

All MLP-based models use ReLU as the hidden-layer activation function, chosen for its computational efficiency and strong empirical performance \citep{Guo2017DeepFM}. Each CVR head applies a sigmoid to its output logit to produce the bucket-specific hazard probability described in Section~\ref{sec:pred}.
The MLP parameters --- one weight matrix and one bias vector per layer --- are learned via backpropagation. Tunable hyperparameters include the number of layers, units per layer, learning rate, dropout rate, and L2 regularization weight; full search ranges are listed in Appendix~\ref{Appendix HPO}.
The two baselines require minimal tuning. The LR has no tunable hyperparameters and the RF is fixed to the settings in Table~\ref{tab:HP_RF}, with the exception of the minimum leaf size, which is tuned via the same nested rolling window procedure as the neural network models (Section~\ref{sec:eval}).

\subsection{Benchmark Models}

\looseness=-1 We benchmark MTL-MHOL against two classical baselines: Logistic Regression (LR), a simple and widely used reference in CVR prediction, and a Random Forest (RF) as a tree-based alternative. Both produce a single CVR estimate and therefore train only on fully mature observations. Additionally, we include the Feedback Shift Importance Weighting model \citep[FSIW,][]{yasui2020} as a competitive delayed feedback baseline, and  the Entire Space Multi-task Model \citep[ESMM,][]{ma2018entire} as a competitive benchmark for addressing data sparsity. Both FSIW and ESMM are implemented with an MLP backbone (see Appendices~\ref{app:fsiw} and~\ref{app:esmm}), and all models share the same input features for consistency. Neural network models use the same embedding-based representation for categorical variables, whereas the classical baselines use one-hot encoding.

\subsection{Evaluation}\label{sec:eval}

We evaluate predictive performance and online learning aptitude via rolling window cross-validation. In each fold, the model is tuned on an inner rolling window covering the outer training set, retrained on the full outer training set, and evaluated against realized outcomes in the test period. For the public dataset, 5 outer folds and 2 inner folds are used; the private data are evaluated on 6 outer folds each including one inner fold due to base rate instability.

\looseness=-1 Performance is reported as mean Negative Log-Likelihood (NLL) and Relative Cross Entropy (RCE) across folds. Lower NLL indicates better calibrated probabilistic predictions; higher RCE indicates a larger improvement in log-loss over a baseline that always predicts the sample mean --- a model with RCE of 37.9 reduces cross-entropy by 37.9\% over this naive predictor.
Additional PR-AUC scores are reported in Appendix~\ref{app:ranking} to separate discrimination from calibration.
Full details on NLL, RCE and the rolling window design are provided in Appendix~\ref{Appendix HPO}.

\subsection{Compute Budget}

All experiments are run on a single 16-core CPU node rd-fleet.4xlarge (128GB) using PyTorch 2.7.0 on Databricks runtime 17.3LTS. Training time per outer fold is approximately 80-89 minutes for the private datasets and 145 minutes for the Criteo dataset, resulting in a total compute budget of approximately 29 CPU-hours across all datasets and folds, including
hyperparameter tuning.

\section{Results} \label{Sect: results}

\subsection{Ablation Study} \label{sec:ablation}

\looseness=-1 
We conduct an ablation study to isolate the performance contribution of each architectural component. The ablation compares four variants: a shifted MLP (single head, no auxiliary task, trained only on fully mature observations), MTL (shifted MLP plus auxiliary task), MHOL (multi-head, no auxiliary task), and the full MTL-MHOL. Comparing shifted MLP with MTL and MHOL isolates the effect of each component individually; comparing MHOL with MTL-MHOL and MTL with MTL-MHOL reveals their incremental contributions when added to the other; and comparing shifted MLP with MTL-MHOL gives the combined effect.

\begin{table}[b!]
\centering
\setlength{\tabcolsep}{0pt}
\caption{\looseness=-1
Mean predictive performance across folds for MTL-MHOL, its ablation variants (MHOL, MTL, MLP), and the ESMM, FSIW, LR and RF benchmarks. Standard deviations across folds are shown in parentheses. Best result in each column in bold. Note that the results for multi-head variants in the private data might be slightly biased (see Appendix~\ref{app:implementation details}).
}
\label{tab:pred_performance}
\begin{tabular}{
@{}
l
@{\hspace{1.5cm}}
>{\centering\arraybackslash}p{1.55cm}
>{\centering\arraybackslash}p{1.55cm}
@{}
>{\centering\arraybackslash}p{1.55cm}
>{\centering\arraybackslash}p{1.55cm}
@{}
>{\centering\arraybackslash}p{1.55cm}
>{\centering\arraybackslash}p{1.55cm}
@{}
}
\toprule
& \multicolumn{2}{c}{Criteo}
& \multicolumn{2}{c}{Customer 1}
& \multicolumn{2}{c}{Customer 2} \\
\cmidrule(lr){2-3}\cmidrule(lr){4-5}\cmidrule(lr){6-7}
& NLL (\(\downarrow\)) & RCE (\(\uparrow\))
& NLL (\(\downarrow\)) & RCE (\(\uparrow\))
& NLL (\(\downarrow\)) & RCE (\(\uparrow\)) \\
\midrule
MTL-MHOL
  & \shortstack{\textbf{0.1215} \\ \textbf{\small (0.0092)}}
  & \shortstack{\textbf{36.704} \\ \textbf{{\small (0.874)}}}
  & \shortstack{\textbf{0.1318} \\ \textbf{{\small (0.0059)}}}
  & \shortstack{\textbf{4.621}  \\ \textbf{{\small (1.804)}}}
  & \shortstack{\textbf{0.0937} \\ \textbf{{\small (0.0060)}}}
  & \shortstack{\textbf{11.962} \\ \textbf{{\small (1.467)}}} \\[4pt]
MHOL
  & \shortstack{0.1222 \\ {\small (0.0094)}}
  & \shortstack{36.282 \\ {\small (0.895)}}
  & \shortstack{0.1330 \\ {\small (0.0059)}}
  & \shortstack{3.752  \\ {\small (1.804)}}
  & \shortstack{0.0949 \\ {\small (0.0057)}}
  & \shortstack{10.821 \\ {\small (0.666)}}  \\[4pt]
MTL
  & \shortstack{0.1350 \\ {\small (0.0052)}}
  & \shortstack{29.499 \\ {\small (2.874)}}
  & \shortstack{0.1331 \\ {\small (0.0047)}}
  & \shortstack{3.733  \\ {\small (0.525)}}
  & \shortstack{0.0946 \\ {\small (0.0058)}}
  & \shortstack{11.107 \\ {\small (1.689)}}  \\[4pt]
MLP
  & \shortstack{0.1362 \\ {\small (0.0065)}}
  & \shortstack{28.880 \\ {\small (2.098)}}
  & \shortstack{0.1334 \\ {\small (0.0079)}}
  & \shortstack{3.486  \\ {\small (1.776)}}
  & \shortstack{0.0953 \\ {\small (0.0061)}}
  & \shortstack{10.437 \\ {\small (1.926)}}  \\[4pt]
FSIW
  & \shortstack{0.1221 \\ {\small (0.0086)}}
  & \shortstack{36.357 \\ {\small (1.025)}}
  & \shortstack{-- }
  & \shortstack{-- }
  & \shortstack{-- }
  & \shortstack{-- } \\[5pt]
ESMM
  & \shortstack{0.1337 \\ {\small (0.0053)}}
  & \shortstack{30.189 \\ {\small (3.017)}}
  & \shortstack{-- }
  & \shortstack{-- }
  & \shortstack{-- }
  & \shortstack{-- } \\[4pt]
RF
  & \shortstack{0.1684 \\ {\small (0.0061)}}
  & \shortstack{12.049 \\ {\small (3.230)}}
  & \shortstack{0.1332 \\ {\small (0.0049)}}
  & \shortstack{3.602  \\ {\small (0.132)}}
  & \shortstack{0.0979 \\ {\small (0.0055)}}
  & \shortstack{7.981  \\ {\small (0.319)}}  \\[4pt]
LR
  & \shortstack{0.1527 \\ {\small (0.0068)}}
  & \shortstack{20.295 \\ {\small (2.565)}}
  & \shortstack{0.1353 \\ {\small (0.0057)}}
  & \shortstack{2.053  \\ {\small (0.865)}}
  & \shortstack{0.0976 \\ {\small (0.0058)}}
  & \shortstack{8.293  \\ {\small (1.051)}}  \\
\bottomrule
\end{tabular}
\end{table}

\begin{table}[b!]
\vskip -0.5em
\centering
\caption{Calibration bias across datasets. $\Delta$ denotes the absolute gap between the per-fold base rate and the average predicted probability. Lower $\Delta$ indicates better calibration.}
\label{tab:cal_deltas}
\begin{tabular}{lcccccc}
  \toprule
  & \multicolumn{2}{c}{Criteo} & \multicolumn{2}{c}{Customer 1} &
  \multicolumn{2}{c}{Customer 2} \\
  \cmidrule(lr){2-3} \cmidrule(lr){4-5} \cmidrule(lr){6-7}
  Model & mean $\Delta$ & Std. dev. & mean $\Delta$ & Std. dev. &
  mean $\Delta$ & Std. dev. \\
  \midrule
  MTL-MHOL & 0.00264 & 0.00179 & 0.00447 & 0.00332 & 0.00280 & 0.00325 \\
  MHOL     & 0.00292 & 0.00186 & 0.00505 & 0.00288 & 0.00478 & 0.00285 \\
  MTL      & 0.03230 & 0.00542 & 0.00543 & 0.00379 & 0.00547 & 0.00392 \\
  MLP      & 0.03426 & 0.00409 & 0.00648 & 0.00376 & 0.00720 & 0.00329 \\
  FSIW     & 0.00474 & 0.00420 &  --     &    --   &   --    &   --    \\
  ESMM     & 0.03264 & 0.00477 &  --     &    --   &   --    &   --    \\
  RF       & 0.03812 & 0.00729 & 0.00143 & 0.00082 & 0.00395 & 0.00086 \\
  \bottomrule
\end{tabular}
\vskip -0.7em
\end{table}

\paragraph{Component ordering.} \looseness=-1 Table~\ref{tab:pred_performance} shows that MTL-MHOL outperforms every reduced variant across all datasets. For Criteo and Customer~1, the hierarchy MLP$<$MTL$<$MHOL$<$MTL-MHOL emerges, with a relative RCE gain of approximately 30\% over the MLP baseline in both cases, indicating that the multi-head component accounts for the larger share of the improvement while the auxiliary task provides a consistent additional gain.
For Customer~2 the ordering of the two intermediate variants reverses to MLP$<$MHOL$<$MTL$<$MTL-MHOL, suggesting that for this customer the auxiliary task contributes more than the additional partially mature observations. The overall RCE gain for Customer 2 is also more modest, at around 15\%, approximately half that of Criteo and Customer~1.

\paragraph{Calibration.} A key mechanism behind the multi-head advantage is calibration. Table~\ref{tab:cal_deltas} reports the average absolute deviation of predicted probabilities from the per-fold base rate. For multi-head models, this calibration error is roughly an order of magnitude smaller than for their single-head counterparts on Criteo, where the large share of partially mature observations in the training set causes single-head models to be systematically trained on a converter-enriched sample, inducing an upward bias in predicted CVR. On the private datasets (Table~\ref{tab:cal_deltas}), delay horizons are short and fewer than 1\% of training observations are partially mature, so this enrichment is negligible and calibration differences across variants are small, with only 0.002 and 0.004 separating the shifted MLP from MTL-MHOL for Customers~1 and 2 respectively. PR-AUC scores in Table~\ref{tab:pr_auc} show that this pattern is restricted to the intercept bias rather than affecting the ability to rank observations, with all NN models performing at a similar level in this regard.

\paragraph{Miscalibration impact.} \looseness=-1 To quantify how much of the single-head deficit is attributable to miscalibration rather than lack of predictive signal, we apply an intercept-only variant of Platt scaling \citep{platt1999probabilistic} to the shifted MLP on Criteo, using the unmasked final training day as a held-out calibration set. This lowers NLL to 0.1216 and raises RCE to 36.694, reducing the gap to MTL-MHOL.
MTL-MHOL achieves comparable performance without any post-hoc correction, avoiding the additional bias that arises when the base rate shifts between the calibration and test periods. The calibrated single-head figures should therefore be read as an optimistic upper bound rather than a practically realizable benchmark.

\paragraph{Complementarity of components.} The monotone improvement pattern across all datasets and the reinforcing pattern in RCE gains support the conclusion that the two components contribute complementary information rather than overlapping signal. The multi-head architecture reduces calibration bias and introduces temporal structure through bucket-specific hazard modeling, while the auxiliary task stabilizes the shared representation via denser supervision. For Customer~2, the slightly worse calibration of MTL relative to MHOL, despite its lower NLL and higher RCE, suggests that the auxiliary task improves discrimination even when it does not reduce intercept bias. 
The PR-AUC results in Table~\ref{tab:pr_auc} support this picture: models with an MTL component consistently rank higher in discrimination across all datasets, with MTL-MHOL-type models leading in every case. Notably, MTL outperforms MHOL on all three datasets in PR-AUC, mirroring the NLL and RCE pattern of Customer 2 and confirming that the auxiliary task provides discriminative signal beyond what the multi-head component alone captures.

\subsection{Benchmark Comparison} \label{sec:benchmark}

\looseness=-1 Table~\ref{tab:pred_performance} reports the predictive performance of MTL-MHOL against the LR, RF, ESMM, and FSIW benchmarks. ESMM and FSIW are evaluated on the Criteo dataset only: the private datasets' short delay horizons render the counterfactual correction largely ineffective, and data limitations hinder a fair translation of the entire space modeling proposition (see Appendix~\ref{app:fsiw} and \ref{app:esmm}).
MTL-MHOL outperforms both basic benchmarks across all datasets, with the largest absolute margins on the Criteo dataset, where the model achieves its strongest improvements in both NLL and RCE.

\looseness=-1 On Criteo, MTL-MHOL achieves 20\% lower NLL and 81\% higher RCE than LR. RF is the weakest classical baseline in its raw form --- MTL-MHOL reduces its NLL by 28\% and roughly triples its RCE. Table~\ref{tab:cal_deltas} shows that this deficit is driven primarily by miscalibration: RF's average absolute deviation from the per-fold base rate is 0.038 against a mean base rate of 4.93\%, with consistently positive per-fold biases indicating systematic overestimation of CVR. This overestimation stems from the stale training set, in which recent converters are included but recent non-conversions have not yet matured, inflating the positive rate that RF conditions on. Post-hoc Platt scaling \citep{platt1999probabilistic} largely corrects this bias: an intercept-only correction lifts RF to an NLL of 0.1535 and RCE of 19.956, close to LR, while adding a slope correction yields an NLL of 0.1414 and RCE of 26.306, exceeding LR and approaching the single-head neural variants. These calibrated figures rely on an offline calibration set and represent an optimistic upper bound. MTL-MHOL, by contrast, is well calibrated without any post-hoc correction.

\looseness=-1 On the private datasets, MTL-MHOL consistently outperforms both benchmarks. For Customer~1, it achieves 2.6\% lower NLL and 125\% higher RCE than LR, and 1.1\% lower NLL and 28.29\% higher RCE than RF; for Customer~2, 4\% lower NLL and 44\% higher RCE than LR, and 4.3\% lower NLL and 49.88\% higher RCE than RF. Unlike on Criteo, RF is well calibrated on the private data (Table~\ref{tab:cal_deltas}), as the short delay horizons leave the stale training set barely enriched with converters and the upward bias in predicted CVR largely disappears. Overall, absolute gains are largest on Criteo; the highest relative RCE improvement over LR occurs for Customer~1, though this reflects a low RCE base rather than a larger absolute gain. MTL-MHOL is preferred across all datasets, with the most substantial benefits on Criteo.

\looseness=-1 
On the two neural benchmarks for the public dataset, MTL-MHOL outperforms ESMM by 21.5\% in RCE and edges FSIW by a modest 1\%. Notably, MHOL alone does not outperform FSIW, trailing it by 0.2\% in RCE --- in contrast to \citet{gao2022multihead}, who benchmarked against an LR-backed FSIW. Hence, with a neural backbone FSIW and MHOL compete at the same level, while the auxiliary task in MTL-MHOL provides the margin. Calibration-wise, FSIW's importance weights largely correct the intercept bias of single-head models, achieving calibration errors roughly an order of magnitude smaller than other single-head variants, though still slightly behind the multi-head models. ESMM, on the other hand, does not have a mechanism to mitigate the intercept bias arising from the matured data restriction, which is reflected in its performance being comparable to the MTL model.

\subsection{Bucket-Level Accuracy}

\begin{figure}[t!]
    \vskip -0.4em
    \centering

    \begin{subfigure}{0.32\textwidth}
        \centering
        \includegraphics[width=\linewidth]{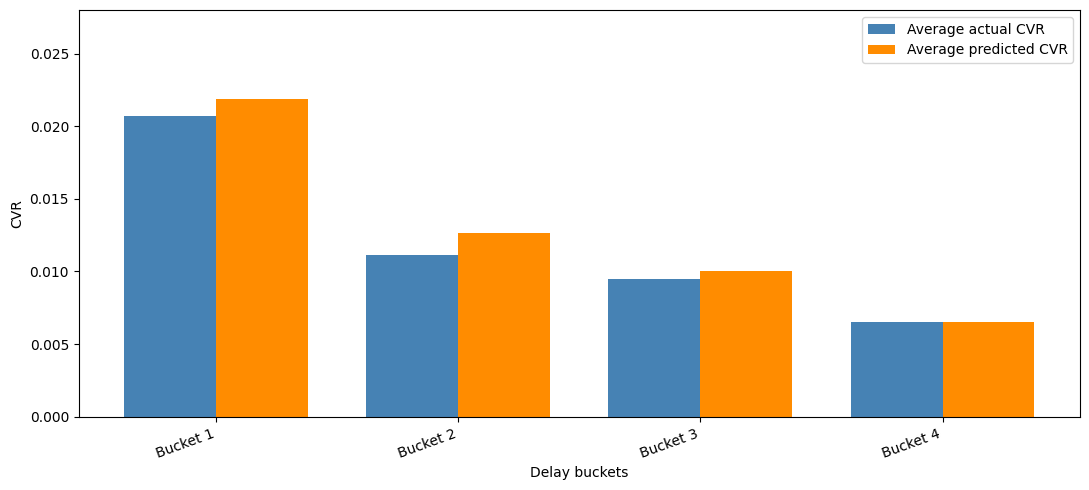}
        \caption{Criteo data}
        \label{subfig:criteo_mh_perf}
    \end{subfigure}
    \hfill
    \begin{subfigure}{0.32\textwidth}
        \centering
        \includegraphics[width=\linewidth]{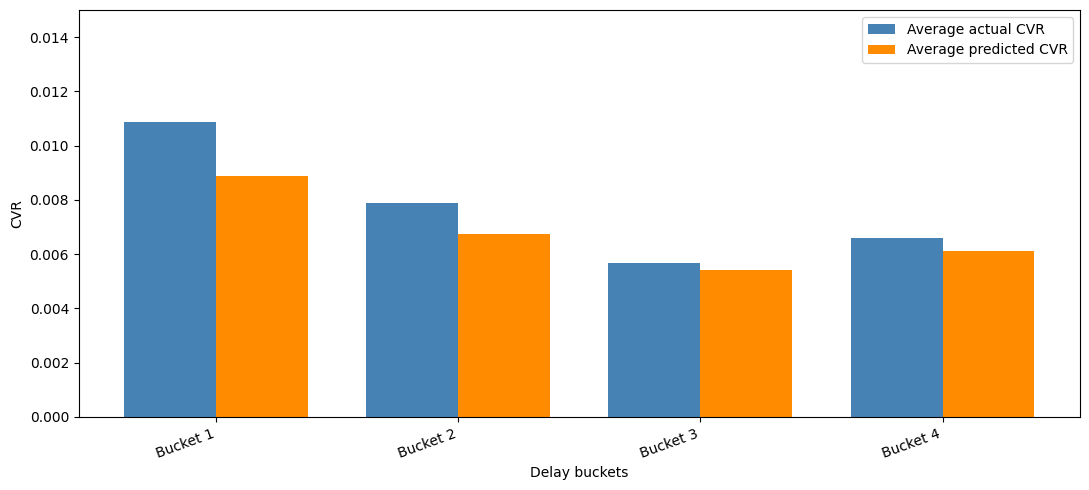}
        \caption{Private data (Customer 1)}
        \label{subfig:bg_mh_perf}
    \end{subfigure}
    \hfill
    \begin{subfigure}{0.32\textwidth}
        \centering
        \includegraphics[width=\linewidth]{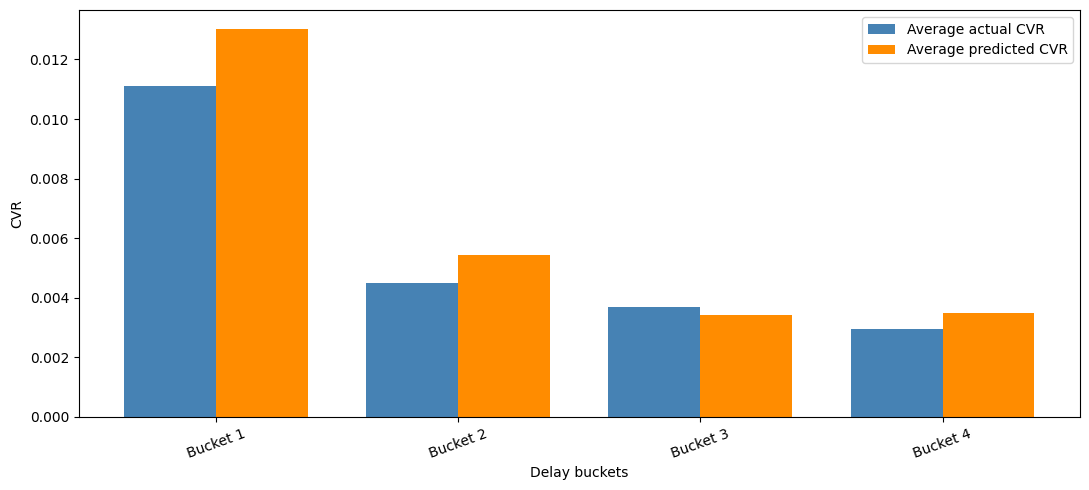}
        \caption{Private data (Customer 2)}
        \label{subfig:customer2_mh_perf}
    \end{subfigure}


    \caption{Predicted CVR of MTL-MHOL compared to actual CVR for each time bucket on all three datasets. Note that the results for the private data might be slightly biased (see Appendix~\ref{app:implementation details}).}
    \label{fig:mh_performance}
    \vskip -0.6em
\end{figure}

\looseness=-1 Figure~\ref{fig:mh_performance} evaluates whether MTL-MHOL predicts not only if a conversion occurs but also when, by comparing predicted against actual CVR for each delay bucket averaged across outer folds. Results are shown for Criteo (Figure~\ref{subfig:criteo_mh_perf}), Customer~1 (Figure~\ref{subfig:bg_mh_perf}), and Customer~2 (Figure~\ref{subfig:customer2_mh_perf}).
In all datasets, the model correctly reproduces the bucket-level hierarchy: bucket 1 consistently shows the highest CVR, and the dataset-specific temporal patterns are accurately captured. For Criteo and Customer~2, predicted CVR decreases monotonically across buckets, matching the observed trend. For Customer~1, CVR decreases through bucket 3 before rising again, and the model successfully tracks this non-monotone pattern as well. This indicates that MTL-MHOL is able to learn different temporal structures across datasets rather than imposing a fixed decay shape.

\looseness=-1 Residual discrepancies between predicted and actual CVR are small throughout. On Criteo, the model shows a slight upward bias across most buckets, with a maximum deviation of 0.002. For Customer~1, the direction reverses to a slight downward bias, with a maximum deviation of 0.003. Neither bias is large enough to affect the bucket-level ordering, and both are consistent with the aggregate calibration results in Table~\ref{tab:cal_deltas}. The largest discrepancy for Customer~2 is 0.002, which is, again, not substantial.
Overall, these results confirm that MTL-MHOL can distinguish between imminent and delayed conversions.

\subsection{Model-Agnostic Evaluation} \label{sec:model-agnostic}

Table~\ref{tab:deepfm} compares the MLP and DeepFM implementations of MTL-MHOL across all datasets. On the private datasets, the DeepFM variant outperforms the MLP variant by up to 10\% in RCE; on Criteo, the MLP yields approximately 3\% higher RCE.

\begin{table}[t!]
    \vskip -0.5em
\centering
\setlength{\tabcolsep}{0pt}
\caption{\looseness=-1
Mean predictive performance across folds for MTL-MHOL with DeepFM and MLP workhorse
models against the LR and RF benchmarks.
Standard deviations across folds are shown in parentheses.
Best result in each column in bold. Note that the results for multi-head variants in the case of the private data might be slightly biased (see Appendix~\ref{app:implementation details}).
}
\label{tab:deepfm}
\makebox[\textwidth][c]{%
\begin{tabular}{
@{}
l
@{\hspace{1.5cm}}
>{\centering\arraybackslash}p{1.55cm}
>{\centering\arraybackslash}p{1.55cm}
@{}
>{\centering\arraybackslash}p{1.55cm}
>{\centering\arraybackslash}p{1.55cm}
@{}
>{\centering\arraybackslash}p{1.55cm}
>{\centering\arraybackslash}p{1.55cm}
@{}
}
\toprule
& \multicolumn{2}{c}{Criteo}
& \multicolumn{2}{c}{Customer 1}
& \multicolumn{2}{c}{Customer 2} \\
\cmidrule(lr){2-3}\cmidrule(lr){4-5}\cmidrule(lr){6-7}
& NLL (\(\downarrow\)) & RCE (\(\uparrow\))
& NLL (\(\downarrow\)) & RCE (\(\uparrow\))
& NLL (\(\downarrow\)) & RCE (\(\uparrow\)) \\
\midrule
MTL-MHOL (DeepFM)
  & \shortstack{0.1235 \\ {\small (0.0084)}}
  & \shortstack{35.592 \\ {\small (1.604)}}
  & \shortstack{\textbf{0.1312} \\ \textbf{{\small (0.0060)}}}
  & \shortstack{\textbf{5.063}  \\ \textbf{{\small (0.542)}}}
  & \shortstack{\textbf{0.0932} \\ \textbf{{\small (0.0054)}}}
  & \shortstack{\textbf{12.409} \\ \textbf{{\small (0.596)}}} \\[5pt]
MTL-MHOL (MLP)
  & \shortstack{\textbf{0.1215} \\ \textbf{{\small (0.0092)}}}
  & \shortstack{\textbf{36.704} \\ \textbf{{\small (0.874)}}}
  & \shortstack{0.1318 \\ {\small (0.0059)}}
  & \shortstack{4.621  \\ {\small (1.804)}}
  & \shortstack{0.0937 \\ {\small (0.0060)}}
  & \shortstack{11.962 \\ {\small (1.467)}} \\[5pt]
RF
  & \shortstack{0.1684 \\ {\small (0.0061)}}
  & \shortstack{12.049 \\ {\small (3.230)}}
  & \shortstack{0.1332 \\ {\small (0.0049)}}
  & \shortstack{3.602  \\ {\small (0.132)}}
  & \shortstack{0.0979 \\ {\small (0.0055)}}
  & \shortstack{7.981  \\ {\small (0.319)}} \\[5pt]
LR
  & \shortstack{0.1527 \\ {\small (0.0068)}}
  & \shortstack{20.295 \\ {\small (2.565)}}
  & \shortstack{0.1353 \\ {\small (0.0057)}}
  & \shortstack{2.053  \\ {\small (0.865)}}
  & \shortstack{0.0976 \\ {\small (0.0058)}}
  & \shortstack{8.293  \\ {\small (1.051)}} \\
\bottomrule
\end{tabular}
}
    \vskip -0.4em
\end{table}

The private data advantage of DeepFM is consistent with its feature space: 16 sparse categorical fields with many low-frequency labels --- such as rarely active countries or infrequent device types --- create exactly the setting where the FM component excels. By estimating latent interaction vectors, it captures informative pairwise combinations, such as a country indicator interacting with a promotional period dummy, and generalizes to feature combinations unobserved during training.

On Criteo, nine of ten categorical features are anonymized, making it difficult to assess whether meaningful pairwise interactions drive conversion. The MLP's advantage suggests that the relevant structure is better captured through higher-order implicit representations than through explicit second-order interactions. Allocating model capacity to an FM component in this setting appears to constrain the higher-order layers without providing compensating signal.

Together, these results confirm that MTL-MHOL is model-agnostic within the neural network family: both backbones consistently outperform the benchmarks, while the relative advantage of each is strongly data-dependent. This reinforces the recommendation to examine the feature space before committing to a backbone architecture, as the optimal choice varies with the sparsity and cardinality of the categorical fields.

\section{Conclusion} \label{Sect: conclusion}

We proposed MTL-MHOL, a framework that jointly addresses delayed feedback and data sparsity in CVR prediction by combining a delay-aware multi-head architecture with hard parameter-sharing multi-task learning. The multi-head component partitions the conversion horizon into mutually exclusive delay buckets and estimates bucket-specific hazard probabilities via a survival-based aggregation, enabling training on partially mature observations. The auxiliary task enriches the sparse primary signal through shared representations, predicting either a binary target or multiple conditional quantiles depending on the available engagement signal. The framework is evaluated on Criteo and two private customer datasets using NLL and RCE under rolling window cross-validation, with MLP and DeepFM backbones.

\paragraph{Comparative performance.} With the MLP backbone, MTL-MHOL outperforms classical single-head baselines (LR and RF trained on fully matured labels), all ablation variants (MLP, MTL, and MHOL), and the entire-space baseline ESMM, and matches the neural delayed-feedback baseline FSIW. The comparative performance with FSIW (1\% RCE improvement for MTL-MHOL) together with the larger gain over ESMM (21.5\% RCE) suggests that delayed feedback and data sparsity are best addressed jointly.

\paragraph{Multi-head architecture.} On Criteo, the multi-head component is the dominant source of performance gains. By exploiting partially mature observations through earlier heads, MTL-MHOL accesses training data entirely unavailable to single-head models, yielding calibration errors an order of magnitude smaller than single-head counterparts without any post-hoc correction. The step from MLP to MHOL accounts for the largest single RCE jump in this dataset. 

\paragraph{Auxiliary task.} The auxiliary task provides consistent complementary gains on top of the multi-head architecture across all datasets, with the step from MHOL to MTL-MHOL yielding further RCE improvements in every case. Its relative contribution is largest where the primary signal is weakest: for Customer~2 the ordering reverses to MLP$<$MHOL$<$MTL$<$MTL-MHOL, confirming that the two components are genuinely complementary rather than redundant.

\paragraph{Temporal structure.} MTL-MHOL produces well-calibrated, temporally structured predictions across all delay buckets, correctly reproducing dataset-specific timing patterns --- a monotonically decreasing CVR for Criteo and a non-monotone pattern for Customer 1. This timing-aware behavior extends the framework's utility to settings where the expected delay of a conversion affects decision-making, such as budget allocation across campaigns with different conversion cycles.

\paragraph{Model agnosticism.} Strong performance under both MLP and
DeepFM backbones confirms that MTL-MHOL is model-agnostic within the
neural network family. DeepFM outperforms MLP on the private datasets,
where sparse categorical fields favor explicit pairwise interaction
modeling, while MLP is preferred on Criteo, where the anonymized
feature space favors higher-order implicit representations.

\section{Future Work} 

Three directions are worth pursuing. First, training on larger datasets with more extensive hyperparameter searches would provide deeper insight into scalability. Second, access to click timestamps would enable more refined CTR--CVR modeling, for example by conditioning conversion prediction explicitly on clicks. Third, applying the framework to settings with longer delay horizons would more directly stress-test the multi-head architecture and is expected to yield gains beyond those observed on the private datasets.

\section{Limitations} \label{sec:limitations}

\paragraph{Delay horizon as the primary performance driver.} \looseness=-1
The larger absolute gains on Criteo relative to the private datasets can primarily be explained by the difference in maximum delay horizon. The multi-head architecture allows MTL-MHOL to train on partially mature observations through heads that mature earlier, whereas single-head benchmarks can only use fully mature observations. This advantage scales with the delay horizon: on the private datasets, the maximum delay is about one day, thus observations mature quickly and fewer than 1\% of training observations are partially mature. On Criteo, the delay window spans 12 days within a 30-day observation period, leaving roughly 50\% of training observations partially mature. MTL-MHOL can exploit all of them, whereas single-head models can only use the subset that has already converted. The larger pool of recent, partially mature observations directly explains the stronger gains on Criteo.

\paragraph{Practical implications for the private setting.}
The short delay horizon of the private datasets reflects a limitation of the current session-level data rather than of the framework itself. The private company employs more sophisticated session-stitching methods that would yield a longer and more realistic delay horizon. In that setting, the share of partially mature observations at training time would increase substantially, and the multi-head architecture is expected to deliver gains comparable to those observed on Criteo.

\paragraph{Base rate instability on private data.} \looseness=-1
Another limiting factor on the private datasets is the considerable variation in base rates across folds. When the conversion rate shifts between training and test periods, the training set is a poor representation of the test set. This increases the risk of overfitting and reduces the effectiveness of hyperparameter tuning. Consistently, Customer~1 has the most stable base rates across folds and also shows the largest relative RCE improvement over LR.

\paragraph{Signal-to-noise ratio.}
The overall RCE level is lower on the private datasets than on Criteo for all models, including LR. This suggests that the private datasets contain less signal relative to noise, which compresses the absolute gains available to any model and reduces the relative advantage of a flexible architecture over a simple baseline that generalizes well under low signal conditions.

\section{Responsible Deployment}

This work may improve conversion-rate prediction under delayed feedback and data sparsity, with potential benefits for online advertising systems through more efficient budget allocation, better calibration, and improved responsiveness to changing user behavior.
At the same time, stronger conversion models can make personalized targeting and profiling more effective, which may raise privacy concerns when user data are used to infer preferences more precisely than users expect. Moreover, attributes or delay patterns differing across user groups can lead to uneven exposure, unequal treatment or the amplification of existing biases through proxy features. Deploying such models may also introduce feedback loops, 
potentially leading to reinforcement of existing biases or to miscalibration of the model.
Any real-world deployment should therefore incorporate fairness and disparate-impact audits, as well as continuous monitoring for distribution shifts, calibration drift, and performance across relevant user groups, with periodic retraining or recalibration to identify and mitigate these privacy, fairness, and feedback-loop risks.

\newpage
\bibliographystyle{tmlr}
\bibliography{bibliography}

\newpage
\appendix

\counterwithin{figure}{section}
\counterwithin{table}{section}
\renewcommand{\thefigure}{\thesection.\arabic{figure}}
\renewcommand{\thetable}{\thesection.\arabic{table}}

\section{Methodology}

\subsection{Auxiliary Task Literature} \label{app:aux}

Multi-task learning was first introduced by \citet{caruana1993multitask}, and assumes that the primary and auxiliary task share useful underlying representations \citep{caruana1993multitask,CombiningTasks, Ruder2017}. When trained jointly, this enables shared lower layers to learn generalizable features and task-specific upper layers to capture task-dependent patterns, thereby acting as a form of inductive bias that can reduce overfitting on the sparse main task \citep{Caruana1997,ZhangYang2017,Ruder2017}. 

Although the literature suggests that auxiliary tasks can enhance learning of complex relationships, there is no general method for selecting tasks without risking negative transfer \citep{BingelSoegaard2017,Ruder2017,ZhangYang2017}. Existing hypotheses on successful task combinations remain highly context-dependent and do not generalize well across domains \citep{CombiningTasks,alonso2017when,BingelSoegaard2017}.  Therefore, it is advisable to carefully base the decision on what auxiliary task to use on a series of theoretical and empirical considerations.
    
\looseness=-1 In the context of conversion prediction, multi-task formulations that jointly model click or other engagement signals with conversion have been shown to alleviate extreme label sparsity and improve conversion performance by letting the rare conversion task leverage denser behavioral supervision \citep{kitada2019conversion,liu2023task,dishi2025practical}. According to \citet{XUE2026104450}, 20 out of 34 studies report improvements over single-task models, while the remaining studies show mixed or inconsistent results. More generally, when the auxiliary variable is observed more frequently than conversion events, this allows the primary task to benefit from a richer and more informative supervision signal during training, improving representation quality for the main task \citep{BingelSoegaard2017, alonso2017when}.

\subsection{Auxiliary Weight Sensitivity Analysis} \label{app:sens_analysis}

Table~\ref{tab:lambda_sensitivity_all} shows that sensitivity to $\lambda_{\text{aux}}$ depends on both the model variant and base rate stability. For MTL-MHOL the relative RCE spread stays below 0.04 on Criteo and widens to 0.042 and 0.087 for Customers~1 and~2, where base rates are less stable (Section~\ref{sec:limitations}), suggesting that tuning on the first fold and fixing thereafter is preferable to a universal default. MTL is markedly more sensitive, with spreads of 0.049, 0.295, and 0.125 across the three datasets. In absolute terms, the worst-to-best $\lambda_{\text{aux}}$ gap for MTL-MHOL corresponds to at most 0.80 RCE points --- small relative to the architectural gains in Table~\ref{tab:pred_performance}.

\begin{table}[h]
\centering
\caption{
Sensitivity of MTL and MTL-MHOL to $\lambda_{\mathrm{aux}}$ across all datasets, with all other hyperparameters held fixed. RCE is normalized to the value at $\lambda_{\mathrm{aux}} = 0.5$ for each model--dataset pair (baseline = 1.000).
}
\label{tab:lambda_sensitivity_all}
\setlength{\tabcolsep}{7pt}
\begin{tabular}{lcccccc}
\toprule
& \multicolumn{2}{c}{Criteo}
& \multicolumn{2}{c}{Customer 1}
& \multicolumn{2}{c}{Customer 2} \\
\cmidrule(lr){2-3}\cmidrule(lr){4-5}\cmidrule(lr){6-7}
& MTL & MTL-MHOL
& MTL & MTL-MHOL
& MTL & MTL-MHOL \\
\midrule
$\lambda_{\text{aux}} = 0.1$
  & 0.967 & 0.969
  & 0.836 & 0.958
  & 0.939 & 1.074 \\
$\lambda_{\text{aux}} = 0.5$
  & 1.000 & 1.000
  & 1.000 & 1.000
  & 1.000 & 1.000 \\
$\lambda_{\text{aux}} = 0.9$
  & 1.016 & 1.003
  & 1.131 & 0.975
  & 1.064 & 0.987 \\
\bottomrule
\end{tabular}
\vskip -0.5em
\end{table}

\subsection{Auxiliary Type Comparison} \label{app:aux_target_type}

\begin{table}[h!]
\centering
\caption{
Auxiliary target type comparison for Customer~2. MTL-MHOL is retrained with a binarized engagement target under identical folds and HPO budget; the continuous (quantile) row is reproduced from Table~\ref{tab:pred_performance}. Standard deviations across folds in parentheses.
}
\label{tab:aux_target_type}
\begin{tabular}{lcc}
\toprule
 & NLL ($\downarrow$) & RCE ($\uparrow$) \\
\midrule
MTL-MHOL (binary)   & 0.0950 (0.0063) & 10.742 (2.056) \\
MTL-MHOL (quantile) & 0.0937 (0.0060) & 11.962 (1.467) \\
\bottomrule
\end{tabular}
\end{table}

Table~\ref{tab:aux_target_type} compares MTL-MHOL with a binarized versus continuous auxiliary target for Customer~2. The binary target labels sessions exceeding the 90th percentile in distinct pages visited as engaged. Even though the performance difference is small, the binary auxiliary task seems to introduce slight negative transfer properties given its performance trailing MHOL's slightly. This ties back to the discussion in Appendix~\ref{app:aux}, where we posit that the auxiliary task is to be chosen ad hoc by the researcher given the possibility for negative transfer and the auxiliary task literature not having established a concrete guideline for selecting the correct auxiliary task in a context-free setting. In this case in particular, the binarization of the continuous task creates an additional dimension of complexity through the need for a cut-off threshold for engaged sessions. This hyperparameter could be the difference between a positive and negative transfer signal being introduced. We therefore advise practitioners to carefully select the auxiliary target based on data availability and expertise, preferably employing continuous auxiliary tasks when possible, given their richer signal.

\subsection{Implementation Details} \label{app:implementation details}

\paragraph{Fair comparison.} All models share the same data preprocessing pipeline; the only divergence occurs at the model input, where neural networks use dense embeddings while LR and RF use one-hot encoded categorical features. Throughout, all models are evaluated on the same folds and evaluation metrics. The full codebase is available at \url{https://github.com/emmavanbreukelen/MTL-MHOL}.

\paragraph{Compute environment.} All experiments run on a single CPU node rd-fleet.4xlarge (128GB, 16 cores) using Databricks runtime 17.3LTS with Apache Spark 4.0.0 and Scala 2.13. Relevant package versions are: torch (2.7.0+cpu), numpy (2.1.3), pandas (2.2.3), scikit-learn (1.6.1), optuna (3.6.1). Further details and packages can be found in the corresponding README.md file.

\paragraph{Training time.} Hyperparameter optimization uses the TPE sampler in Optuna with 32 cold-start trials per inner fold, reduced to 12 when the previous outer fold's optimal configuration performs approximately equally well on the current fold's inner folds. On Criteo, the full pipeline across 5 outer folds takes approximately 724 minutes, amounting to approximately 145 minutes per outer fold. On the private datasets, total runtime is approximately 484 minutes for Customer~1 and 534 minutes for Customer~2 across 6 outer folds. These figures can vary substantially depending on the number of folds, the fraction running the full 32-trial versus the reduced 12-trial budget, and the number of observations in the dataset. Runtime can be further reduced by increasing the tolerance threshold for activating the reduced trial budget, or by using a GPU-based compute cluster.

\paragraph{Correction of the risk-set restriction.} An earlier version of our implementation omitted the at-risk indicator $r_{ib}$ (Equation~\ref{eq:riskset}) from the primary loss in Equation~\ref{Lbce}, so that each head $b$ was also trained on observations that had already converted in an earlier bucket. For Criteo, we re-ran the full pipeline with the corrected code, and the corrected and original results differ by at most $2.1\%$ in NLL and $3.3\%$ in RCE.  For the private datasets, we did not re-run the pipeline and instead show below that the resulting bias is smaller than on Criteo.

Without $r_{ib}$, earlier converters enter the training population of head $b$ with label $y_{ib} = 0$, since at most one bucket per observation can equal one. As noted in Section~\ref{sec:loss}, head $b$ then converges to the marginal $\Pr(y_{ib} = 1 \mid \mathcal{X}_i)$ rather than the hazard $h_{ib}$ of Equation~\ref{eq:hazard}, and the two are related by $\Pr(y_{ib} = 1 \mid \mathcal{X}_i) = h_{ib} \Pr(y_{i1} = \dots = y_{i(b-1)} = 0 \mid \mathcal{X}_i)$. The estimated hazard is therefore scaled down by the probability of not having converted before bucket $b$, so that its relative bias equals the probability of having converted before bucket $b$. This bias is zero for the first bucket and largest for the last. Averaged over observations, it equals the base rate multiplied by the share of conversions falling in buckets $1$ to $b-1$. For the last bucket, this gives $4.3\%$ for Criteo versus $2.4\%$ and $2.0\%$ for Customers 1 and 2 (Tables~\ref{tab:summarystats}, \ref{tab:bucket_distribution} and \ref{tab:criteo_bucket_distribution}).

These per-bucket biases largely cancel in the overall conversion probability $\hat{p}_i$ on which NLL and RCE are evaluated. Because the buckets are mutually exclusive, the marginals $\Pr(y_{ib} = 1 \mid \mathcal{X}_i)$ sum exactly to $\Pr(y_i = 1 \mid \mathcal{X}_i)$. The only remaining error comes from inserting marginals instead of hazards into the product of Equation~\ref{eq:phat}. This product differs from the sum of its terms only through cross-products of bucket-level conversion probabilities. Evaluated at the population-level conversion probabilities, the uncorrected implementation underestimates $\hat{p}_i$ by approximately $1.7\%$ for Criteo and $1.2\%$ and $0.8\%$ for Customers 1 and 2 in relative terms.

The bias on the private datasets is therefore roughly half to two-thirds of that on Criteo, both per bucket and after aggregation. Given the limited difference between the corrected and original results on Criteo, we expect the uncorrected private-data results to be affected even less.

\section{Data Analysis}
\label{Appendix Data}

\begin{table}[h!]
\centering
\caption{Summary statistics for all datasets. Values for the downsampled dataset used for modeling are reported first, with corresponding values for the full dataset in parentheses.}
\label{tab:summarystats}
\begin{tabular}{lccc}
\hline
\textbf{Statistic} & \textbf{Criteo} & \textbf{Customer 1} & \textbf{Customer 2} \\
\hline
Number of observations  & 247,199 (16,468,027) & 174,971 (8,262,943) & 215,031 (4,097,427) \\
Number of conversions   & 11,993 (806,196)     & 5,350 (144,305)     & 4,922 (53,087)      \\
Conversion rate         & 4.93\% (4.90\%)      & 3.13\% (1.75\%)     & 2.34\% (1.30\%)     \\
Numerical predictors    & 4                    & 5                   & 5                   \\
Categorical predictors  & 10                   & 16                  & 16                  \\
Observation period      & 30 days             & Jun 2024--Aug 2025  & May 2024--Aug 2025  \\
\hline
\end{tabular}
\end{table}

\subsection{Private Data} \label{Appendix: private data}

The private data consist of three anonymized dataframes: the
\emph{Sessions Dataframe} (website behavior and contextual information),
the \emph{Marketing Spend Dataframe} (marketing exposure and spend), and
the \emph{Events Dataframe} (conversions and revenue). A $uid$ represents
a single browser cookie; a \emph{session} is an uninterrupted website
visit by a given $uid$; and an \emph{attribution session} aggregates
multiple sessions of the same $uid$ not separated by long periods of
inactivity. The three dataframes are merged into a single session-level
dataset, where each session is the observation unit.

\paragraph{Downsampling.} Sessions with at most one distinct page visit
and a duration of five seconds or less are removed as uninformative.
Additional random downsampling is then applied to obtain a computationally
feasible dataset size.

\paragraph{Feature engineering.} Two groups of features are
constructed. First, within-user dynamics are captured by ordering
sessions chronologically per $uid$, enabling session-index, rolling
averages of historical session duration and distinct pages visited,
and indicators for previous conversions. Second, temporal context
variables are added: a Black Friday indicator, three-hour time-of-day
categories, and day-of-week, week-of-month, and month indicators.

\paragraph{Preprocessing.} Strictly positive numerical variables are
log-transformed; variables that may contain zeros use a log-one-plus
transformation. All numerical variables are then standardized using
training-set parameters only. Missing numerical values are imputed to
zero (the post-standardization mean) following \citet{che2018recurrent},
with a binary missingness indicator added per variable.
Since the variables here lack the time-series structure of \citet{che2018recurrent}, we impute the post-standardization mean directly. The model can learn the informational content of missingness through the associated indicator flags.
Categorical variables are integer-encoded, with missing values and training-unseen categories mapped to explicit unknown indices to prevent implicit learning from categories unavailable at prediction time.

\paragraph{Descriptive statistics.}\label{Appendix Descriptive Stats}
Tables~\ref{stats_cust1} and \ref{stats_cust3} summarize the variables used for modeling.
(Note that not all variables from the original dataset are retained; marketing spend and revenue are excluded as they cannot be linked consistently.)

Several categorical variables exhibit high cardinality (device and campaign identifiers) or strong concentration in a single category (country, Black Friday). Numerical variables are substantially right-skewed, reflecting large variation in user behavior.

\begin{figure}[h!]
        \centering
        \begin{subfigure}[t]{0.49\textwidth}
            \centering
            \includegraphics[width=0.76\textwidth]{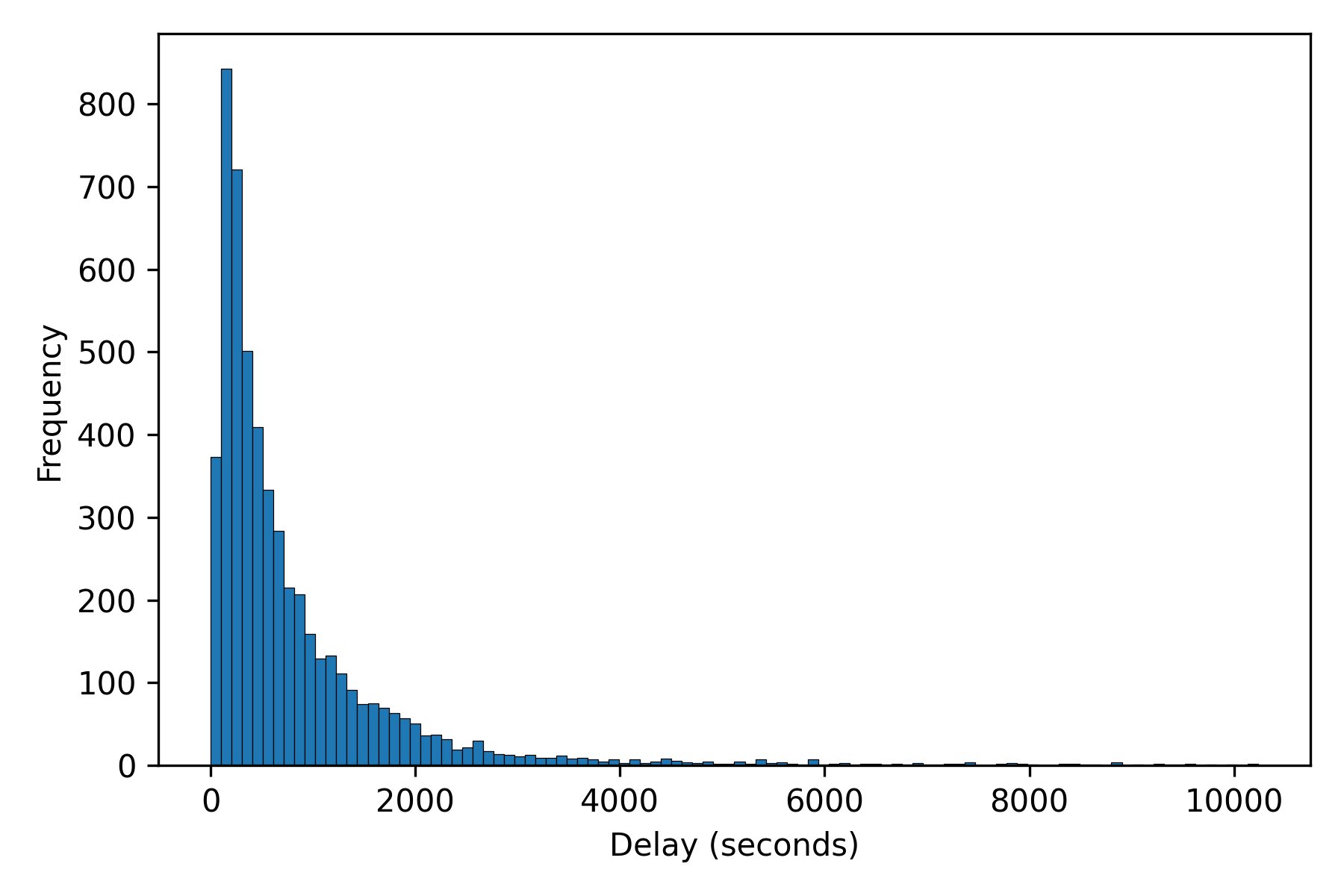}
            \label{delay_c2_97.5}
        \end{subfigure}
        \hfill
        \begin{subfigure}[t]{0.49\textwidth}
            \centering
            \includegraphics[width=0.76\textwidth]{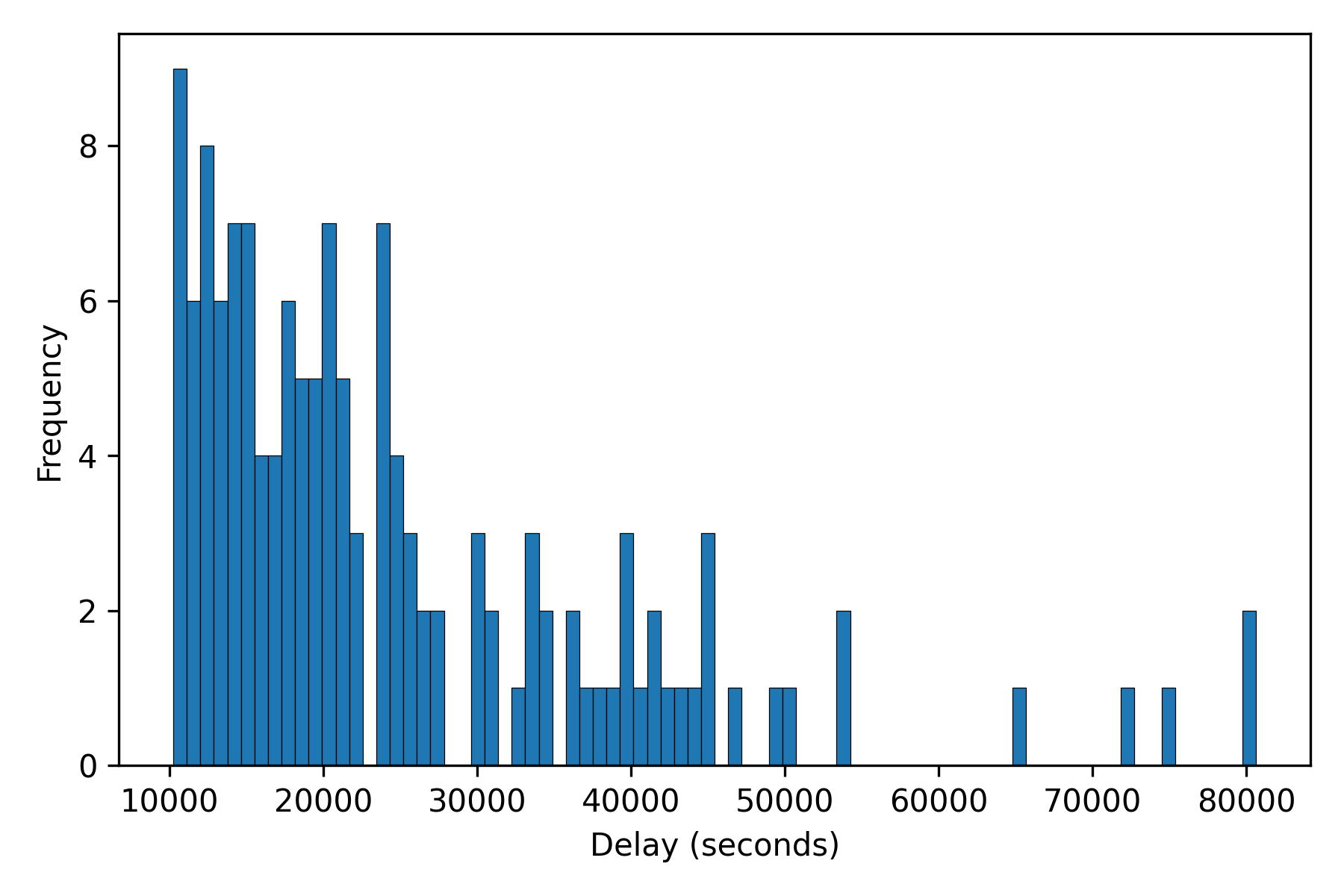}
            \label{delay_c2_2.5}
        \end{subfigure}
        \vspace{-4mm}
        \caption{Conversion Delays Customer 1, up to the 97.5th percentile (a) and the upper $2.5\%$ tail (b), after truncation of negative delays and downsampling.}
        \label{Delay_c1}
    \end{figure}

\begin{figure}[h!]
        \centering
        \begin{subfigure}[t]{0.48\textwidth}
            \centering
            \includegraphics[width=0.78\textwidth]{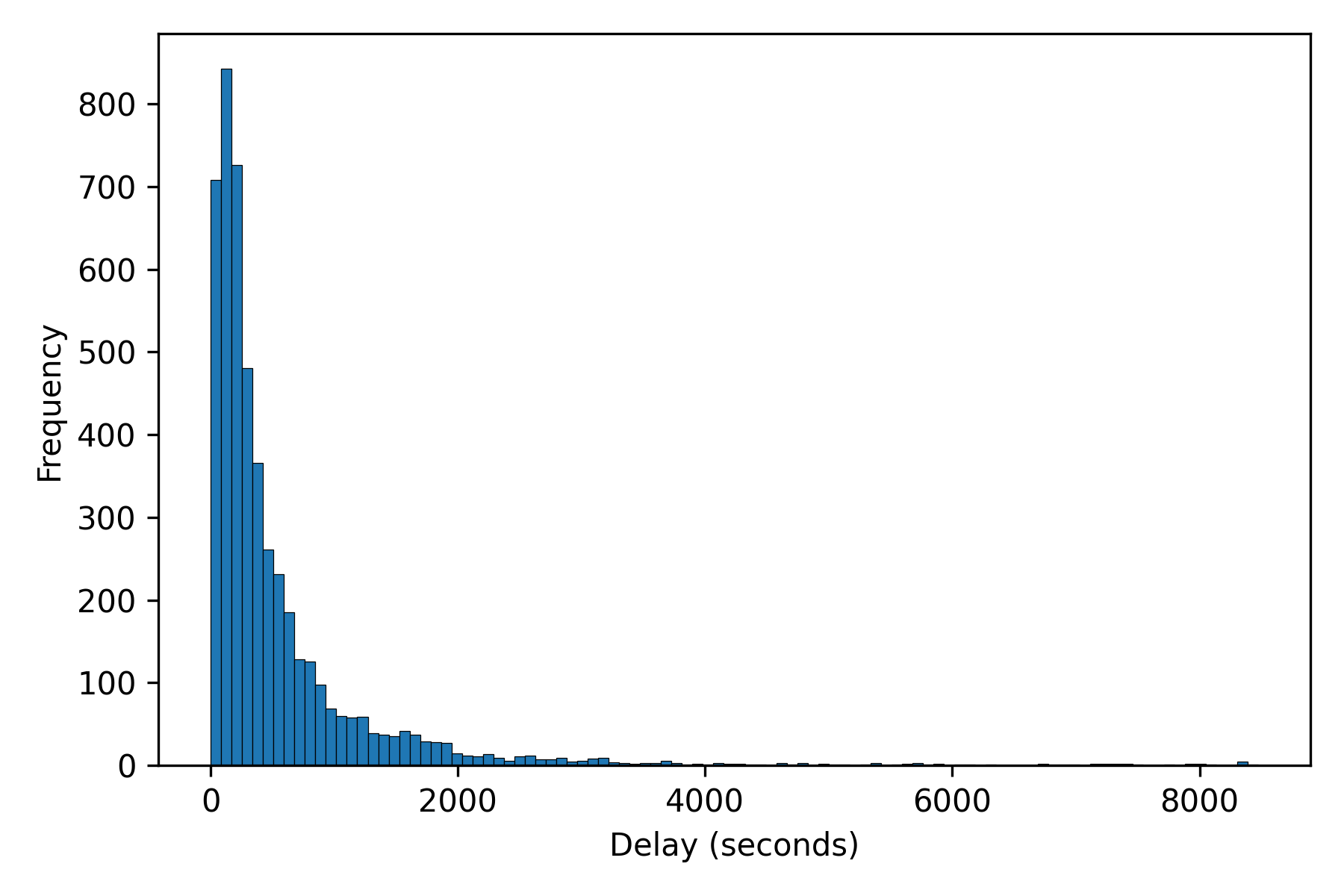}
            \label{delay_c2_97.5}
        \end{subfigure}
        \hfill
        \begin{subfigure}[t]{0.49\textwidth}
            \centering
            \includegraphics[width=0.78\textwidth]{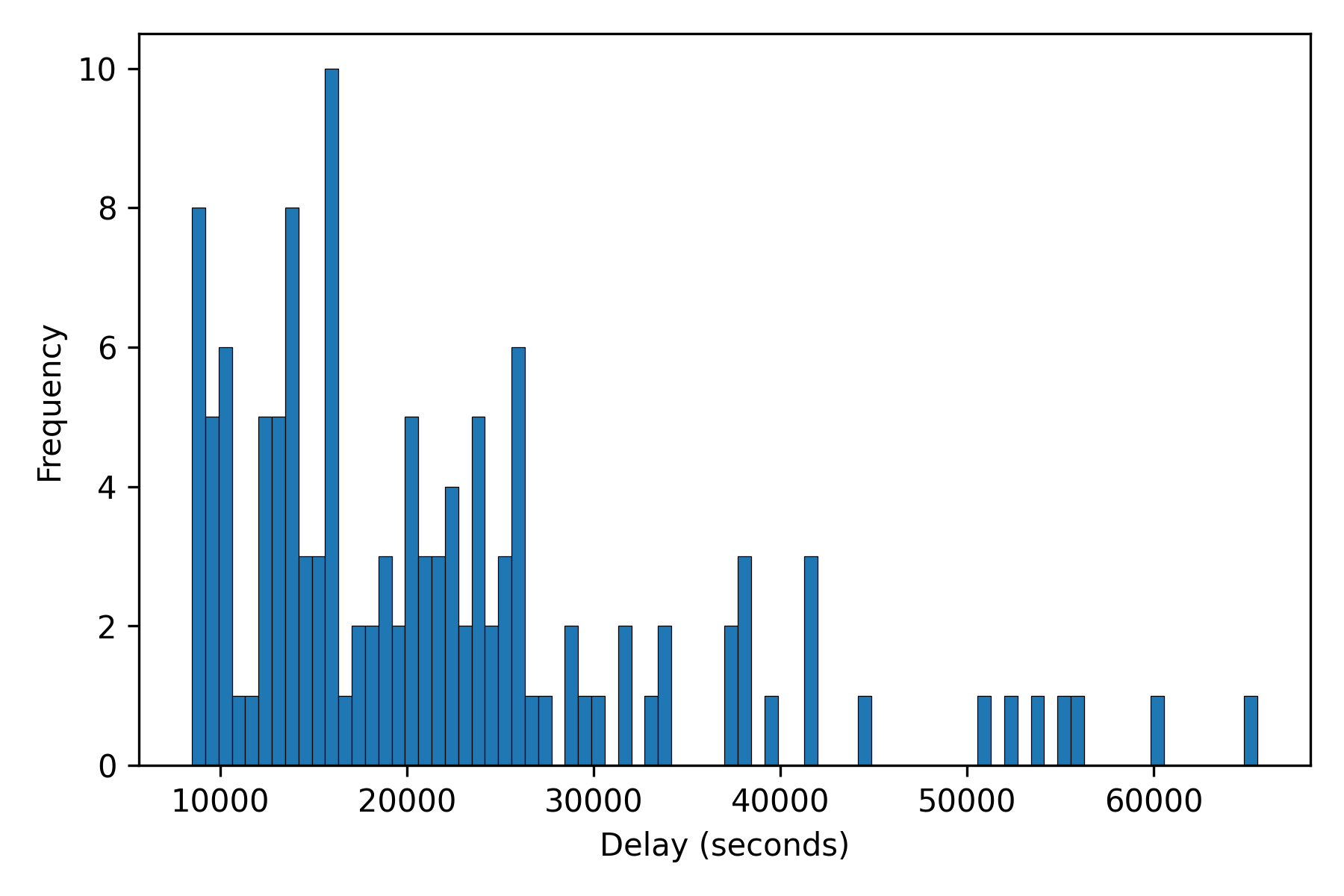}
            \label{delay_c2_2.5}
        \end{subfigure}
        \vspace{-4mm}
        \caption{Conversion Delays Customer 2, up to the 97.5th percentile (a) and the upper $2.5\%$ tail (b), after truncation of negative delays and downsampling.}
        \label{Delay_c3}
    \end{figure}

\paragraph{Delay.} \label{Appendix Delay private}
Conversion delay is defined as the time between session start and the last recorded event timestamp. A small number of sessions with negative delays, likely due to logging errors, are truncated to zero. Figures~\ref{Delay_c1} and \ref{Delay_c3} show the resulting delay distributions for both customers: strongly right-skewed with a small number of long delays and a maximum of approximately 18--24 hours.

\paragraph{Anonymization.}

All identifiers, monetary values and exposure values are hashed or anonymized by the private company to avoid linkage to real individuals or entities. 

    \begin{table}[h!]
    \vskip -1em
        \centering
        \caption{\looseness=-1 Retained model features after downsampling and before standardizing for Customer~1.}
        \small
        \begin{tabular}{lcc}
        \hline
        \textbf{Variable} & \textbf{Number of Categories} & \textbf{Largest Share Category (\%)} \\
        \hline
        \multicolumn{3}{l}{\textit{Categorical Variables}} \\
        \hline
        country & 98 & 88.12 \\
        device\_info\_1\_transformed & 21 & 48.73 \\
        device\_info\_2\_transformed & 630 & 39.18 \\
        device\_info\_3\_transformed & 712 & 36.21 \\
        browser\_info\_1\_transformed & 24 & 27.21 \\
        os\_info\_1\_transformed & 8 & 41.39 \\
        source\_transformed & 291 & 60.86 \\
        medium\_transformed & 22 & 55.37 \\
        campaign\_id\_transformed & 111 & 30.84 \\
        adset\_id\_transformed & 418 & 30.84 \\
        ad\_id\_transformed & 1330 & 30.84 \\
        weekday\_categorical & 7 & 17.79 \\
        month & 12 & 12.19 \\
        week\_of\_month & 4 & 33.89 \\
        time\_of\_day\_categorical & 8 & 20.43 \\
        black\_friday & 2 & 96.04 \\
        \hline
        \end{tabular}
        
        \vspace{0.4cm}  
        
        \begin{tabular}{lccccc}
        \hline
        \textbf{Variable} & \textbf{Min} & \textbf{Max} & \textbf{Mean} & \textbf{Median} & \textbf{SD} \\
        \hline
        \multicolumn{6}{l}{\textit{Numerical Variables}} \\
        \hline
        seconds\_since\_last\_session & 0 & 31,325,381 & 377,536.869 & 16,648 & 1,298,471.051 \\
        session\_number & 1 & 3,768 & 4.771 & 1 & 31.869 \\
        avg\_prev\_pages\_visited & 1 & 280 & 5.66 & 3.167 & 7.529 \\
        avg\_prev\_session\_duration & 0 & 9,792 & 243.624 & 104.5 & 399.303 \\
        n\_previous\_conversions & 0 & 23 & 0.032 & 0 & 0.219 \\
        \hline
        \end{tabular}
        \label{stats_cust1}
    \end{table}

\enlargethispage{\baselineskip}

    \begin{table}[h!]
        \centering
        \caption{Retained model features after downsampling and before standardizing for Customer~2.}
        \small
        \begin{tabular}{lcc}
        \hline
        \textbf{Variable} & \textbf{Number of Categories} & \textbf{Largest Share Category (\%)} \\
        \hline
        \multicolumn{3}{l}{\textit{Categorical Variables}} \\
        \hline
        country & 147 & 69.33 \\
        device\_info\_1\_transformed & 27 & 71.11 \\
        device\_info\_2\_transformed & 911 & 59.87 \\
        device\_info\_3\_transformed & 1006 & 43.4 \\
        browser\_info\_1\_transformed & 29 & 36.08 \\
        os\_info\_1\_transformed & 7 & 60.84 \\
        source\_transformed & 247 & 52.78 \\
        medium\_transformed & 20 & 38.18 \\
        campaign\_id\_transformed & 101 & 38.69 \\
        adset\_id\_transformed & 263 & 38.69 \\
        ad\_id\_transformed & 1843 & 38.69 \\
        weekday\_categorical & 7 & 15.68 \\
        month & 12 & 16.04 \\
        week\_of\_month & 4 & 31.12 \\
        time\_of\_day\_categorical & 8 & 18.69 \\
        black\_friday & 2 & 98.45 \\
        \hline
        \end{tabular}
        
        \vspace{0.4cm}  
        
        \begin{tabular}{lccccc}
        \hline
        \textbf{Variable} & \textbf{Min} & \textbf{Max} & \textbf{Mean} & \textbf{Median} & \textbf{SD} \\
        \hline
        \multicolumn{6}{l}{\textit{Numerical Variables}} \\
        \hline
        seconds\_since\_last\_session & 0 & 34,308,807 & 580,157.594 & 47,394 & 1,983,867.389 \\
        session\_number & 1 & 5,981 & 8.609 & 1 & 141.879 \\
        avg\_prev\_pages\_visited & 1 & 101 & 3.108 & 2.409 & 2.583 \\
        avg\_prev\_session\_duration & 0 & 5,351 & 176.947 & 62 & 306.038 \\
        n\_previous\_conversions & 0 & 74 & 0.062 & 0 & 1.448 \\
        \hline
        \end{tabular}
        \label{stats_cust3}
    \end{table}

\enlargethispage{\baselineskip}    

\subsection{Criteo Data} \label{Appendix: criteo data}

\looseness=-1 Each observation corresponds to an impression --- exposure to an online advertisement --- after which the individual may or may not click. A conversion can follow either a click or, less commonly, direct engagement with the ad without clicking. The observation period covers one month, significantly shorter than the private datasets.

\paragraph{Downsampling.} 
Conversions are first downsampled to a maximum delay window of $H = 12$ days, since the original 30-day window would cover the full dataset and leave almost no fully mature observations available for single-head training. Non-conversions are downsampled by the same percentage to preserve the original conversion rate, after which additional random downsampling is applied for computational feasibility.

\paragraph{Feature engineering.} \looseness=-1
Since the observation period covers only one month, time-specific variables such as Black Friday and week-of-month indicators are omitted as they are unlikely to yield additional predictive signal. The remaining features are limited to those required for training and evaluation, as reported in Table~\ref{stats_criteo}.

\paragraph{Preprocessing.} 
Preprocessing follows the same pipeline as for the private data (Section~\ref{Appendix: private data}): numerical variables are log-transformed and standardized using training-set parameters only, missing values are imputed to zero with binary missingness indicators added per variable, and categorical variables are integer-encoded with unseen categories mapped to an explicit unknown index. Timestamps are converted to datetime format to enable accurate time-based computations.

\paragraph{Descriptive statistics.}\label{Appendix Criteo descriptive stats}
\looseness=-1 Table~\ref{stats_criteo} summarizes the variables used for modeling.
The dataset contains four numerical and ten categorical predictors, of
which nine categorical features are anonymized. Categorical variables
vary substantially in cardinality and concentration: cat1 and cat8 have
few categories while cat3 and cat7 exhibit very high cardinality. Cat4
is heavily imbalanced with its largest category accounting for 92.77\%
of observations, whereas cat3 and campaign are more evenly distributed,
with largest category shares of 2.01\% and 2.71\%. Among the numerical
variables, \texttt{time\_since\_last\_impression} and
\texttt{time\_since\_last\_click} show large ranges and high standard
deviations, while \texttt{n\_previous\_conversions} and
\texttt{impression\_number} are more moderate. The conversion rate is
4.93\% after downsampling, closely matching the 4.90\% of the full
dataset.

\begin{figure}[h!]
        \centering
        \begin{subfigure}[t]{0.49\textwidth}
            \centering
            \includegraphics[width=\textwidth]{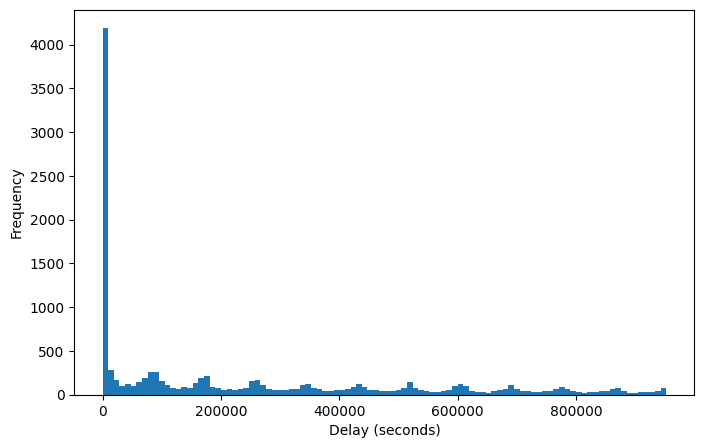}
            \label{delay_criteo_97.5}
        \end{subfigure}
        \hfill
        \begin{subfigure}[t]{0.49\textwidth}
            \centering
            \includegraphics[width=\textwidth]{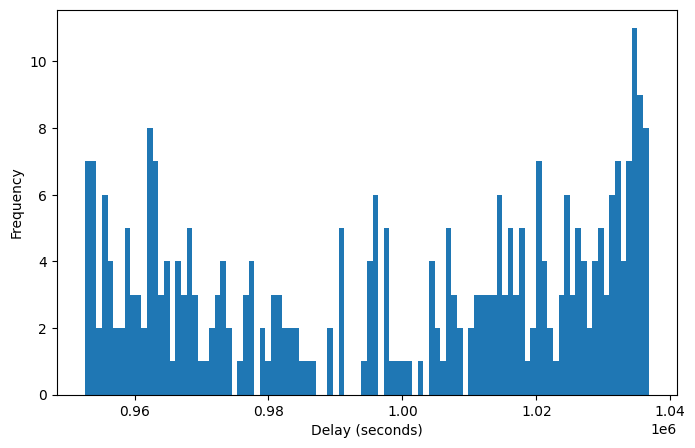}
            \label{delay_criteo_2.5}
        \end{subfigure}
        \vspace{-7mm}
        \caption{\looseness=-1 Conversion Delays Criteo, up to the 97.5th percentile (a), the upper $2.5\%$ tail (b) after downsampling.}
        \label{Delay_criteo}
    \end{figure}

\paragraph{Delay.} \label{Appendix Delay Criteo}
Figure~\ref{Delay_criteo} shows the distribution of conversion delays, defined as the time in seconds from the first impression to conversion. The distribution is strongly right-skewed, with the large majority of conversions occurring within one day of initial exposure. Beyond this range, delays are sparse and spread across the full 12-day horizon, as illustrated by the upper 2.5\% tail in panel~(b).

 \begin{table}[t!]
    \centering
\caption{Descriptive statistics of retained model features after downsampling and before standardizing for the Criteo data.}
    \small
\begin{tabular}{lcc}
\hline
\textbf{Variable} & \textbf{Number of Categories} & \textbf{Largest Share Category (\%)} \\
\hline
\multicolumn{3}{l}{\textit{Categorical Variables}} \\
\hline
campaign & 674 & 2.71 \\
cat1 & 9 & 44.43 \\
cat2 & 60 & 52.66 \\
cat3 & 1720 & 2.01 \\
cat4 & 15 & 92.77 \\
cat5 & 51 & 30.79 \\
cat6 & 30 & 44.27 \\
cat7 & 13115 & 13.85 \\
cat8 & 11 & 59.63 \\
cat9 & 30 & 53.25 \\
\hline
\end{tabular}
    
\vspace{0.4cm}
    
\begin{tabular}{lccccc}
\hline
\textbf{Variable} & \textbf{Min} & \textbf{Max} & \textbf{Mean} & \textbf{Median} & \textbf{SD} \\
\hline
\multicolumn{6}{l}{\textit{Numerical Variables}} \\
\hline
time\_since\_last\_impression & 0 & 2,619,072 & 287,388.097 & 123,908 & 400,459.512 \\
time\_since\_last\_click & 0 & 2,591,995 & 270,685.446 & 0 & 526,826.543 \\
n\_previous\_conversions & 0 & 249 & 0.427 & 0 & 2.983 \\
impression\_number & 1 & 865 & 4.479 & 2 & 8.656 \\
\hline
\end{tabular}
    \label{stats_criteo}
    \end{table}

\subsection{Conversion Distribution Across Delay Buckets} \label{Appendix Bucket construction}

For the private data, bucket 1 corresponds to delays between 0 and 300 seconds (5 minutes), bucket 2 between 300 and 600 seconds (10 minutes), bucket 3 between 600 and 1200 seconds (20 minutes), and bucket 4 between 1200 seconds and $H$. This maximum conversion delay $H$ is 80,651 seconds (22.40 hours) for Customer 1 and 65,550 seconds (18.21 hours) for Customer 2.

\looseness=-1 The Criteo dataset has a longer, more realistic maximum delay horizon of $H = 12$ days.
Most conversions occur within the first day, therefore the first bucket covers delays between 0 and 1 day. The remaining buckets are chosen in a simple and intuitive way such that they still contain enough observations and roughly comparable percentages of the remaining conversions. Bucket 1 corresponds to delays between 0 and 1 day, bucket 2 between 1 and 4 days, bucket 3 between 4 and 8 days, and bucket 4 between 8 days and $H$.

\begin{table}[h!]
\centering
\caption{Distribution of number of conversions across delay buckets for each customer of the private data. Percentages represent the share of total conversions within each dataset.}
\label{tab:bucket_distribution}
\begin{tabular}{lcccc}
\toprule
Bucket & Customer 1  & Customer 2 \\
\midrule
0--300 sec      & 1809 (33.8\%) & 2538 (51.6\%) \\
300--600 sec    & 1322 (24.7\%) & 1017 (20.7\%) \\
600--1200 sec   & 1030 (19.3\%) & 725 (14.7\%) \\
1200 sec -- $H$ & 1189 (22.2\%) & 642 (13.0\%) \\
\bottomrule
\end{tabular}
\end{table}

\begin{table}[h!]
\centering
\caption{Distribution of number of conversions across delay buckets for the Criteo dataset. Percentages represent the share of total conversions.}
\label{tab:criteo_bucket_distribution}
\begin{tabular}{lc}
\toprule
Bucket & Conversions \\
\midrule
0--1 day   & 5466 (45.6\%) \\
1--4 days  & 2773 (23.1\%) \\
4--8 days  & 2247 (18.7\%) \\
8--12 days & 1507 (12.6\%) \\
\bottomrule
\end{tabular}
\end{table}

\section{Evaluation}\label{Appendix HPO}
\subsection{Evaluation Metrics}

\looseness=-1 The models are evaluated on two probability-based evaluation metrics to capture their performance. First, we evaluate the NLL, which measures how well the predicted probabilities match the actual matured outcomes.
For a binary outcome \(y_{i}\in\{0,1\}\) and predicted probability \(\hat{p}_{i}\) of overall conversion, the mean NLL is given by
    \begin{equation}
        \mathrm{NLL} \;=\; -\frac{1}{N}\sum_{i=1}^{N}\Bigl[\,y_{i}\ln\!\left(\hat{p}_{i}\right) + (1-y_{i})\ln\!\left(1-\hat{p}_{i}\right)\Bigr],
    \end{equation}
where $N$ denotes the total number of observations in the test or evaluation set. 
This expression corresponds to a binary cross‑entropy loss \(\mathcal{L}_{\mathrm{CE}}\) (see Equation \ref{Lbce}) for fully observed aggregated probabilities. Lower values therefore indicate better probabilistic predictions, and minimizing NLL is equivalent to minimizing the primary loss function. Here, lower values indicate better calibrated probabilistic predictions.     
Furthermore, we compute the RCE, which evaluates how much better our model is compared to a simple benchmark, in our case predicting the average outcome. Let \(\mathrm{NLL}_{\text{model}}\) be the NLL of the evaluated model, and let \(\mathrm{NLL}_{\text{baseline}}\) be the NLL of a simple baseline that always predicts the sample mean \(\bar{y}=\tfrac{1}{N}\sum_{i=1}^{N}y_{i}\) (with \(\hat{p}_{i}=\bar{y}\) for all \(i\)). We define RCE as the relative improvement of the model over the baseline:
    \begin{equation}
        \mathrm{RCE} \;=\; 1 \;-\; \frac{\mathrm{NLL}_{\text{model}}}{\mathrm{NLL}_{\text{baseline}}}
    \;=\; \frac{\mathrm{NLL}_{\text{baseline}}-\mathrm{NLL}_{\text{model}}}{\mathrm{NLL}_{\text{baseline}}}.
    \end{equation}
    Here, positive values indicate that the model reduces cross‑entropy relative to the baseline. Moreover, both RCE and NLL penalize overconfident wrong predictions.
    
\subsection{Rolling Window Cross-Validation}

The models are evaluated using a rolling window cross-validation setup to ensure a realistic and time-sensitive evaluation. As opposed to random cross-validation, this approach adheres to the chronological order of the data and therefore prevents data leakages from future observations into the training process.

The procedure consists of an outer and an inner loop. In the outer loop, the model is trained on a fixed window and tested on the period directly after. This window then moves forward step by step, creating multiple overlapping folds. Within each outer fold, hyperparameters are tuned using an inner cross-validation loop on the training data only. This inner loop also follows a rolling window structure, where the data are split into smaller training and validation periods while maintaining the time order to prevent leakage.

    For the private datasets, in each outer fold, the model is trained on eight consecutive months and tested on the following two months. After each outer fold, the window shifts forward by one month and the procedure is repeated. This approach results in overlapping folds where the training window maintains a constant length while gradually moving forward through time. This process begins at the earliest month available in the dataset and continues until there is no longer enough data to complete a full test window.

    For the Criteo dataset, online learning can be mimicked even better. Here, in each outer fold, the model is trained on the previous 25 days and tested on the following day, with a step size of one day. This reflects a setting where models are updated daily as new data becomes available.
    
    Within each outer fold, hyperparameter tuning is performed using an inner fold cross-validation procedure, where only the training observations of the corresponding outer fold are used. This training period is further divided into smaller training and validation splits for the inner folds, while again preserving the time-based order of the data. 
    For the private datasets, we use six months for training and two months for validation for the inner folds, with a step size of one month. For the Criteo dataset, we use 23 days for training and one day for validation, with a step size of one day. 
    
    The evaluation procedure results in six outer folds for the private datasets and five outer folds for the Criteo dataset, each performing inner folds.
    This nested rolling window design ensures that the performance evaluation is strictly based on past information at each step and therefore prevents data leakage from future observations. Moreover, this setup closely replicates real-world conditions, where models are trained and updated in real time to make accurate predictions.

\subsection{Hyperparameter Optimization}

\looseness=-1 To optimize the performance of our model, we implement hyperparameter tuning using the Tree-structured Parzen Estimator (TPE) sampler in Optuna \citep{watanabe2023tree,akiba2019optuna}. By learning from past trials, it ensures an efficient tuning process and allows us to find good configurations with fewer evaluations. We complement our TPE approach by implementing the Enqueue function, which forces each fold to use the best trial from the previous fold as a prior. If in the new outer fold, the old fold's optimal hyperparameters perform similarly well over all inner folds, then the number of trials is reduced. This makes the tuning process more efficient.
    
The parameters we tune, together with their search spaces and sampling strategies, are displayed in Table \ref{tab:AppendixHP}. 
The neural-network architecture of MTL-MHOL is optimized by tuning batch size, dropout rate, learning rate, L2 weight decay, the hidden layer sizes, embedding dimensions, and the number of epochs. These choices are standard practice in optimizing the performance of a neural network model. Additionally, we tune the number of quantiles in the auxiliary task as well as the weight of the auxiliary task to the total loss function $\lambda_{\text{aux}}$. The latter not only optimizes performance, but also gives us insights into the effectiveness of complementing our multi-head model with an engagement measure to improve CVR prediction.

For each inner fold in our rolling window cross-validation approach, hyperparameter tuning is performed, where by default 32 trials are executed to evaluate the optimal hyperparameters for that inner fold. In case of the enqueued hyperparameters performing strongly on the new fold, too, the trials are reduced to twelve. 
Then, the best-performing configuration of hyperparameters across all inner folds is used to train the model on the full outer fold training window before evaluating it on the outer test window. Performance is evaluated on each outer test fold, and the final reported metrics are the average across all outer folds.
    
We choose RCE as the objective of our hyperparameter tuning approach. 
By maximizing RCE, we ensure the model improves meaningfully upon a simple benchmark while providing well-calibrated predictions.

\begin{table}[h!]
\centering
\caption{Hyperparameter search space used during TPE tuning with Optuna.}
\begin{tabular}{llll}
\toprule
\textbf{Hyperparameter} & \textbf{Type} & \textbf{Search Space} & \textbf{Sampling} \\
\midrule
Batch Size & Categorical & $\{256, 512, 1024, 2048\}$ & Uniform \\
Dropout Rate & Continuous & $[0.0, 0.5]$ & Uniform \\
Learning Rate & Continuous & $[10^{-4}, 5\times10^{-3}]$ & Log-uniform \\
L2 Weight Decay & Continuous & $[10^{-7}, 10^{-3}]$ & Log-uniform \\
Hidden Layer Sizes & Categorical & 
$\{(256,256),(512,512),(256,128)$,\\ & & $(512,256), (128,128,128),$ \\ & & $(512,256,128),(256,128,128)\}$ & Uniform \\
Embedding Dimensions & Categorical & $\{4, 8, 16, 32\}$ & Uniform \\
Number of Epochs & Categorical & $\{10, 11, 12, ..., 25\}$ & Uniform \\
Number of Quantiles & Categorical & $\{3,5,7\}$ & Uniform \\
Auxiliary Loss Weight & Continuous & $[0.05, 1.5]$ & Log-uniform \\
\bottomrule
\end{tabular}
\label{tab:AppendixHP}
\end{table}

The type of search space is determined by the nature of the hyperparameter. We employ categorical spaces in case only a predefined set of discrete options is meaningful and we use continuous spaces when the parameter can take any value within the defined range. Moreover, we apply uniform sampling when all values within the range are equally likely, whereas log-uniform sampling is used when the parameter spans several orders of magnitude and smaller values should be explored more thoroughly.

For the FSIW benchmark, we employ all the same defaults for the auxiliary LightGBM models as specified in \citet{yasui2020}. For the MLP backbone, the same hyperparameter optimization logic is employed as discussed above. The Hyperparameters for the Random Forest benchmark model are given in Table \ref{tab:HP_RF}.

\begin{table}[h!]
\centering
\caption{Hyperparameters (and search space) used for RF.}
\begin{tabular}{llll}
\toprule
\textbf{Hyperparameter}  & \textbf{Setting}  \\
\midrule
Number of Trees &  $500$  \\
Maximum depth  & None \\
Minimum Samples per Split &  $2$ \\
Minimum Samples per Leaf &  $\{50, 100, 150, 200, 250\}$  \\
Maximum Features Considered per Split & square root of total features \\
Maximum Proportion of Samples in each Tree & $0.7$\\
\bottomrule
\end{tabular}
\label{tab:HP_RF}
\end{table}

\newpage
\section{Ranking} \label{app:ranking}

\begin{table}[h!]
\centering
\caption{Mean PR-AUC across folds for MTL-MHOL, its ablation variants (MHOL, MTL, MLP), DeepFM, and the LR, RF, and FSIW benchmarks. FSIW is evaluated only on the Criteo dataset.}
\label{tab:pr_auc}
\begin{tabular}{lccc}
\toprule
Model & Criteo & Customer 1 & Customer 2 \\
\midrule
MTL-MHOL & 0.465 & 0.071 & 0.074 \\
MHOL     & 0.443 & 0.067 & 0.072 \\
MTL      & 0.465 & 0.071 & 0.074 \\
MLP      & 0.451 & 0.067 & 0.071 \\
DeepFM   & 0.448 & 0.074 & 0.076 \\
RF       & 0.304 & 0.066 & 0.069 \\
LR       & 0.422 & 0.069 & 0.061 \\
FSIW     & 0.458 & --    & --    \\
ESMM     & 0.460 & --    & --    \\
\bottomrule
\end{tabular}
\end{table}

\section{FSIW Benchmark}
\label{app:fsiw}

This appendix describes the Feedback Shift Importance Weighting (FSIW) approach of \citet{yasui2020}, included as a single-head neural benchmark that addresses delayed feedback through loss reweighting.

\paragraph{Method.} For observation $i$ with mature label $y_i$ and training cutoff $\varphi$, define the label observed at cutoff as $\tilde{y}_i(\varphi) = \mathbbm{1}(y_i = 1 \land \text{the conversion is recorded by } \varphi)$. Since $\tilde{y}_i(\varphi) \leq y_i$ always, training on $\tilde{y}_i(\varphi)$ biases a model toward underpredicting conversion --- a discrepancy \citet{yasui2020} term a feedback shift. FSIW corrects this by weighting the cross-entropy loss with $\rho_i$, defined as
\begin{align*}
    \rho_i &= \tfrac{1}{P(\zeta_i=1 \mid y_i=1,\,\mathcal{X}_i)}
    && \text{if } \tilde{y}_i(\varphi)=1, \\
    \rho_i &= 1 -
    \tfrac{P(\zeta_i=0,\,y_i=1 \mid \mathcal{X}_i)}
          {P(\tilde{y}_i(\varphi)=0 \mid \mathcal{X}_i)}
    && \text{if } \tilde{y}_i(\varphi)=0,
\end{align*}
where $\zeta_i \in \{0,1\}$ indicates whether $i$ is correctly labeled at $\varphi$. This weighting yields a consistent estimator of the loss under the mature label distribution.

\paragraph{Estimation.} Since both probabilities involve $y_i$ itself, they cannot be estimated directly. \citet{yasui2020} instead introduce a counterfactual deadline $\kappa$ and cutoff $\varphi' = \varphi - \kappa$. Observations starting before $\varphi'$ have, under a sufficiently large $\kappa$, relatively mature outcomes by the real cutoff $\varphi$, allowing two auxiliary datasets to be constructed in which the true label at $\varphi'$ is known. Two LightGBM models \citep{ke2017lightgbm} are trained on these datasets using elapsed time relative to $\varphi'$ and $\mathcal{X}_i$ as inputs; at prediction time, elapsed time relative to the real cutoff $\varphi$ is used instead.

\paragraph{Implementation.} We implement FSIW using the same MLP workhorse as Section~\ref{sec:mlp}, configured as a single head with no auxiliary task, so that FSIW and the shifted MLP ablation differ only in loss weighting. The model minimizes
\begin{equation*}
    \mathcal{L}_{\text{FSIW}} = -\frac{1}{N}\sum_{i=1}^N \rho_i
    \Big[\tilde{y}_i(\varphi)\log(\hat{p}_i) +
    (1-\tilde{y}_i(\varphi))\log(1-\hat{p}_i)\Big].
\end{equation*}
Like MHOL and MTL-MHOL, FSIW trains on all observations with start time before $\varphi$, including immature ones, allowing it to draw on the same partially mature data as our multi-head models. This isolates the benefit of our specific mechanisms over the more basic benefit of using fresh data at all.

The counterfactual deadline $\kappa$ trades off assumption validity against data recency: a longer $\kappa$ makes the full-resolution assumption more accurate but discards more recent training data. \citet{gao2022multihead} set $\kappa$ to five days on a thirty-day delay window; to preserve this proportion we set $\kappa$ to two days for Criteo, given the twelve-day maximum delay window.

\paragraph{Criteo restriction.} We restrict the FSIW benchmark to the Criteo dataset for two reasons. First, scaling $\kappa$ proportionally to the private datasets' maximum delay window of under one day would yield a counterfactual window of only a few hours, leaving a negligibly small sub-population from which to estimate $\rho_i$. Fitting two auxiliary LightGBM models per fold under these conditions would add substantial overhead while correcting almost nothing.

Second, and more fundamentally, the correction is only informative where mislabeling at $\varphi$ is non-negligible. Since $\rho_i \to 1$ as $P(\zeta_i=1 \mid y_i=1, \mathcal{X}_i) \to 1$, FSIW adds value only when a meaningful share of training observations are incorrectly labeled. Table~\ref{tab:cal_deltas} shows that single-head mature-only variants (MLP, MTL, RF) exhibit calibration errors an order of magnitude smaller on the private datasets than on Criteo (mean $\Delta$ of 0.001--0.007 versus 0.032--0.038), confirming that mislabeling from partial maturity is negligible on the private data but substantial on Criteo. This is also consistent with the findings of \citet{yasui2020}, who observe significant FSIW gains only for their seven-day campaign, with no significant improvement at one or three days, attributing the effect to delay being less influential at short horizons. Our private datasets, with maximum delays under one day, fall below even that threshold.

\section{ESMM Benchmark}
\label{app:esmm}

This appendix describes the Entire Space Multi-Task Model (ESMM) of \citet{ma2018entire}, included as a benchmark that addresses data sparsity and sample selection bias through entire-space multi-task learning.

\paragraph{Method.} Conventional CVR models are trained only on clicked impressions but applied at inference time to all impressions, a mismatch \citet{ma2018entire} term sample selection bias. Writing $y_i^{\text{aux}}$ for the click indicator and $y_i$ for the conversion label, the quantity of interest is $p(y_i=1 \mid y_i^{\text{aux}}=1, \mathcal{X}_i)$, which conventional models estimate on the clicked subpopulation only. ESMM instead factorizes the joint probability as
\begin{equation*}
    p(y_i^{\text{aux}}=1,\, y_i=1 \mid \mathcal{X}_i)
    =
    p(y_i^{\text{aux}}=1 \mid \mathcal{X}_i)
    \times
    p(y_i=1 \mid y_i^{\text{aux}}=1,\, \mathcal{X}_i),
\end{equation*}
and estimates both terms over the full impression population, removing the conditioning that causes sample selection bias.

\paragraph{Implementation.} We implement ESMM following \citet{ma2018entire} as closely as our setting permits. A single shared embedding table maps categorical inputs to dense vectors, which are passed into two separate MLP towers --- one producing a predicted click probability $\hat{p}_i^{\text{ctr}}$ and one producing a predicted conditional conversion probability $\hat{p}_i^{\text{cvr}}$ --- with the same architecture as the workhorse model of Section~\ref{sec:workhorse}. Sharing only the embedding table allows the conversion tower to benefit from representations learned from the denser click signal while keeping the remaining tower parameters independent, as in the original architecture. The overall predicted conversion probability reported in Table~\ref{tab:pred_performance} is $\hat{p}_i = \hat{p}_i^{\text{ctr}} \hat{p}_i^{\text{cvr}}$, and the model minimizes
\begin{equation*}
    \mathcal{L}_{\text{ESMM}} =
    -\frac{1}{N}\sum_{i=1}^N
    \Big[y_i^{\text{aux}}\log(\hat{p}_i^{\text{ctr}})
    + (1-y_i^{\text{aux}})\log(1-\hat{p}_i^{\text{ctr}})\Big]
    -\frac{1}{N}\sum_{i=1}^N
    \Big[y_i\log(\hat{p}_i)
    + (1-y_i)\log(1-\hat{p}_i)\Big],
\end{equation*}
matching the loss of \citet{ma2018entire} without additional weighting. Since ESMM has no mechanism for handling delayed feedback, it is trained on the same population of fully matured observations as the LR, RF, and shifted MLP benchmarks, rather than the entire population used by MHOL, MTL-MHOL, and FSIW.

\paragraph{Restriction to Criteo.} We restrict ESMM to the Criteo dataset for two reasons. First, the factorization underlying ESMM requires every conversion to be preceded by a click: a converted impression with $y_i^{\text{aux}}=0$ has no valid decomposition into a click probability and a conditional conversion probability. This assumption does not hold on the private datasets, where a converted session need not include a click on the advertisement itself. Second, on Criteo we verify the assumption directly and find that every training fold contains zero conversions with $y_i^{\text{aux}}=0$, so $\hat{p}_i$ coincides exactly with the conversion probability evaluated against the realized outcome in every fold and no adjustment to the evaluation is required.

\section{DeepFM Implementation}\label{app:deepFM}

Following \citet{Guo2017DeepFM}, we use a shared embedding architecture that combines an FM component for first- and second-order effects with a feed-forward network for higher-order effects. In the original DeepFM formulation for CTR prediction, both parts share the same raw input and the same latent feature embeddings, and the final click probability is obtained from the sum of the FM and Deep Neural Network logits \citep{Guo2017DeepFM}. In the multi-task extension of \citet{yildirim2021bideepfm}, the same shared network is equipped with multiple output heads and optimized by a weighted sum of task losses, where the task-specific parameters are confined to the final layer. 

Relative to \citet{Guo2017DeepFM}, our implementation preserves the shared-embedding DeepFM structure but generalizes the final prediction layer to task-specific heads. Relative to \citet{yildirim2021bideepfm}, it retains the multi-objective hard parameter-sharing idea while replacing the dual-sigmoid setup with a more general family of single-head, multi-head, binary-auxiliary, and quantile-auxiliary objectives.

\begin{figure}[b!]
    \centering
    \includegraphics[width=0.45\linewidth, trim=10 40 10 70, clip]{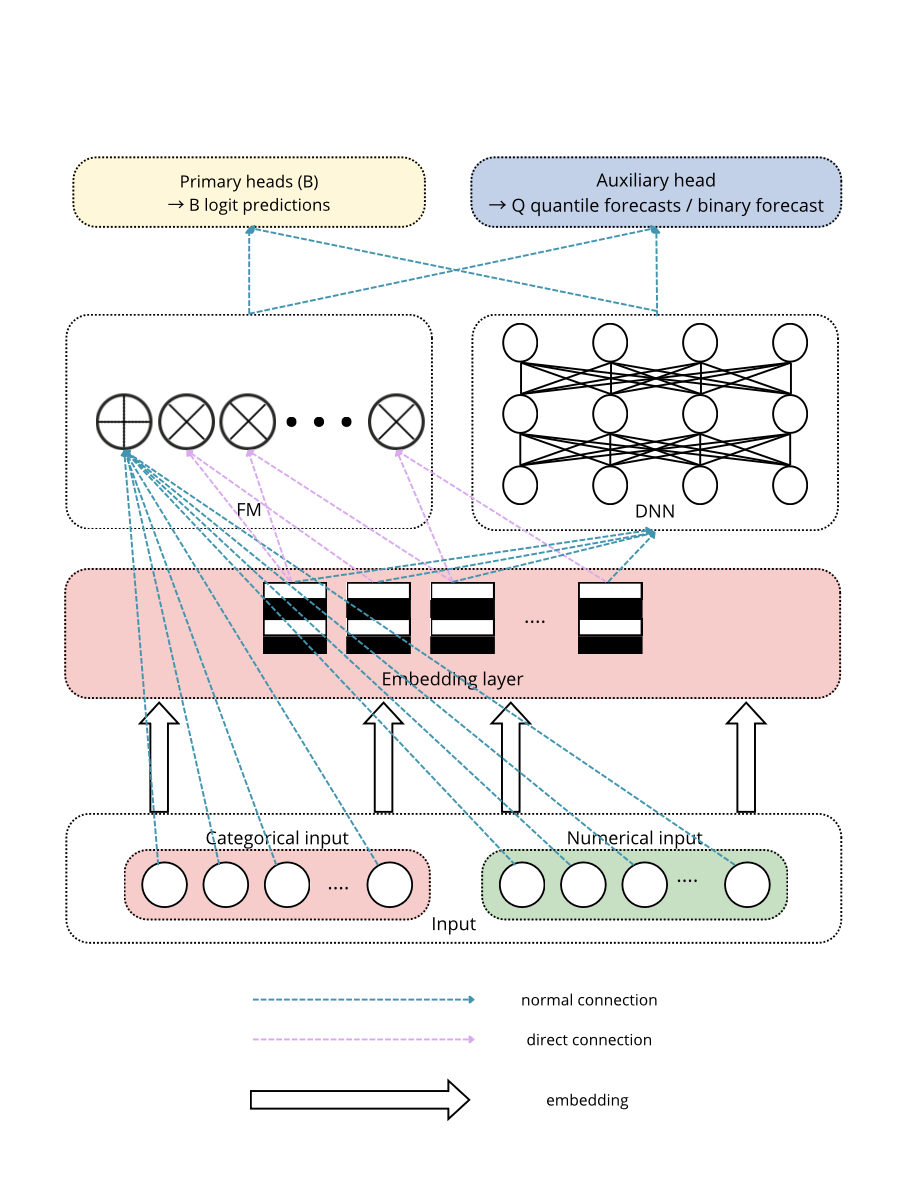}
    \caption{The DeepFM architecture. Blue arrows denote learned weight connections; pink arrows denote fixed weight-1 (direct) connections; thick white arrows denote embeddings. $\oplus$ is summation, $\otimes$ is a dot-product, and empty circles in the MLP denote nonlinear activations (ReLU in this implementation).}
    \label{fig:deepfm}
\end{figure}

\paragraph{Input representation.} Let observation $(\mathbf{x}_i, y_i)$ consist of $F_c$ categorical and $F_n$ numerical fields, with $\mathbf{x}_i = [x_{i,1},\dots,x_{i,F}]$ and $F = F_c + F_n$. For categorical field $j \in \{1,\dots,F_c\}$, let $x_{i,j}^{(c)} \in \{0,\dots,S_j-1\}$ denote its encoded level, where index $0$ is reserved for unseen values at evaluation time. The model learns a first-order scalar weight look-up table $\mathbf{W}_j^{(c)} \in \mathbb{R}^{S_j \times 1}$, giving $w_{i,j}^{(c)} = \mathbf{W}_j^{(c)}[x_{i,j}^{(c)}] \in \mathbb{R}$, and a second-order look-up table $\mathbf{V}_j^{(c)} \in \mathbb{R}^{S_j \times d}$, giving a $d$-dimensional embedding $\mathbf{e}_{i,j}^{(c)} = \mathbf{V}_j^{(c)}[x_{i,j}^{(c)}] \in \mathbb{R}^d$. For numerical field $j \in \{1,\dots,F_n\}$, the scalar $x_{i,j}^{(n)} \in \mathbb{R}$ is preprocessed as described in Section~\ref{Appendix: private data} and embedded multiplicatively as
\begin{equation*}
    \mathbf{e}_{i,j}^{(n)} = x_{i,j}^{(n)} \mathbf{v}_j^{(n)},
    \qquad \mathbf{v}_j^{(n)} \in \mathbb{R}^d,
\end{equation*}
rather than discretized or one-hot encoded as in \citet{Guo2017DeepFM} and \citet{yildirim2021bideepfm}. This avoids rescaling latent vectors by raw numeric values in the FM component without any theoretical alteration. A first-order scalar weight $w_j^{(n)} \in \mathbb{R}$ is also learned per numerical field.

\paragraph{Components.} Collecting all field embeddings $\mathbf{e}_{i,1},\dots,\mathbf{e}_{i,F} \in \mathbb{R}^d$, the FM component is
\begin{equation}
    s_{\mathrm{FM}}(\mathbf{x}_i)
    = b
    + \sum_{j=1}^{F_c} w_{i,j}^{(c)}
    + \sum_{j=1}^{F_n} w_j^{(n)} x_{i,j}^{(n)}
    + \sum_{r=1}^{F} \sum_{s=r+1}^{F}
    \langle \mathbf{e}_{i,r}, \mathbf{e}_{i,s} \rangle,
\end{equation}
where $\langle \cdot, \cdot \rangle$ is the dot product and $b$ is the learned bias. This matches the standard DeepFM/FM decomposition into linear and pairwise interaction terms. The deep component flattens the same embeddings and passes them through an MLP with $R$ output heads:
\begin{equation*}
    \boldsymbol{\eta}(\mathbf{x}_i)
    = \mathrm{MLP}(\mathrm{vec}(\mathbf{e}_{i,1},\dots,\mathbf{e}_{i,F}))
    \in \mathbb{R}^R,
\end{equation*}
following the shared-embedding principle of DeepFM.

\paragraph{Task-specific heads.} Rather than summing one FM logit and one DNN logit as in \citet{Guo2017DeepFM}, we form a shared scalar $s_{\mathrm{FM}}(\mathbf{x}_i)$ and a shared hidden representation $\boldsymbol{\eta}(\mathbf{x}_i)$, then apply task-specific linear maps.

For the primary task with $B$ heads:
\begin{equation*}
    \mathbf{z}^{\mathrm{prim}}(\mathbf{x}_i)
    = \mathbf{A}_{\mathrm{FM}}^{\mathrm{prim}} s_{\mathrm{FM}}(\mathbf{x}_i)
    + \mathbf{A}_{\mathrm{DNN}}^{\mathrm{prim}} \boldsymbol{\eta}(\mathbf{x}_i)
    + \mathbf{b}^{\mathrm{prim}} \in \mathbb{R}^B.
\end{equation*}
When $B=1$ this reduces to a standard single-logit binary model; when $B>1$ each component $z_b^{\mathrm{prim}}(\mathbf{x}_i)$ is a bucket-specific logit and the conversion probability is aggregated as
\begin{equation*}
    \hat{p}(\mathbf{x}_i)
    = 1 - \prod_{b=1}^{B}
    \left(1 - \sigma\!\left(z_b^{\mathrm{prim}}(\mathbf{x}_i)\right)\right),
\end{equation*}
with the per-bucket training loss given in Equation~\ref{Lbce}.

\looseness=-1 For a binary auxiliary target the model outputs:
\begin{equation*}
    z^{\mathrm{aux}}(\mathbf{x}_i)
    = a_{\mathrm{FM}} s_{\mathrm{FM}}(\mathbf{x}_i)
    + \mathbf{a}_{\mathrm{DNN}}^\top \boldsymbol{\eta}(\mathbf{x}_i)
    + b_{\mathrm{aux}},
\end{equation*}
and optimizes a masked binary cross-entropy loss (Equation~\ref{Lbce_aux}). For a quantile auxiliary target with levels $\tau_1 < \cdots < \tau_Q$, the model outputs $\mathbf{q}(\mathbf{x}_i) \in \mathbb{R}^Q$ with
\begin{equation*}
    \mathbf{q}(\mathbf{x}_i)
    = \mathbf{A}_{\mathrm{FM}}^{\mathrm{aux}} s_{\mathrm{FM}}(\mathbf{x}_i)
    + \mathbf{A}_{\mathrm{DNN}}^{\mathrm{aux}} \boldsymbol{\eta}(\mathbf{x}_i)
    + \mathbf{b}^{\mathrm{aux}},
\end{equation*}
and minimizes a masked pinball loss (Equation~\ref{eq_pinball}) with a monotonicity penalty enforcing $q_1(\mathbf{x}_i) \leq \cdots \leq q_Q(\mathbf{x}_i)$, 
mirroring the MLP objective of Section~\ref{sec:loss}.

\end{document}